\documentclass[journal]{IEEEtran}
\ifCLASSINFOpdf
\else
\fi
\usepackage{graphicx}
\usepackage{bigstrut}
\usepackage{multirow}
\usepackage{subcaption}
\usepackage{tabularx}
\usepackage{color}
\usepackage{url}
\usepackage{balance}
\usepackage{wrapfig}
\usepackage{lipsum}
\usepackage{bigstrut}
\usepackage{multirow}
\usepackage{dblfloatfix}
\usepackage[table]{xcolor}

\usepackage{bm} 		%
\usepackage{amsmath}
\usepackage{amssymb}

\begin{document}
\title{Smartphone Multi-modal Biometric Authentication: Database and Evaluation}
\author{Raghavendra Ramachandra,  Martin Stokkenes, Amir Mohammadi, Sushma Venkatesh, Kiran Raja, Pankaj Wasnik, Eric Poiret, S\'{e}bastien Marcel, Christoph Busch
\thanks{Raghavendra Ramachandra, Martin Stokkenes, Sushma Venkatesh, Kiran Raja, Pankaj Wasnik and Christoph Busch are with Norwegian Biometrics Laboratory,  Norwegian University of Science and Technology (NTNU), Norway. e-mail: (raghavendra.ramachandra@ntnu.no).}%
\thanks{Amir Mohammadi and Sebastien Marcel are with Idiap Research Institute, Martigny, Switzerland. e-mail: (\{amir.mohammadi,marcel\}@idiap.ch). }%
\thanks{Eric Poiret is with IDEMIA, France.}%
\thanks{Manuscript received september 19, 2019; revised August 26, 2015.}}
\markboth{Journal of \LaTeX\ Class Files,~Vol.~14, No.~8, August~2015}%
{Shell \MakeLowercase{\textit{et al.}}: Bare Demo of IEEEtran.cls for IEEE Journals}
\maketitle

\begin{abstract}
Biometric-based verification is widely employed on the smartphones for various applications, including financial transactions. In this work, we present a new multimodal  biometric dataset (face, voice, and periocular) acquired using a smartphone. The new dataset is comprised of 150 subjects that are captured in six different sessions reflecting real-life scenarios of smartphone assisted authentication. One of the unique features of this dataset is that it is collected in four different geographic locations representing a diverse population and ethnicity. Additionally, we also present a multimodal Presentation Attack (PA) or spoofing dataset using a low-cost Presentation Attack Instrument (PAI) such as print and electronic display attacks. The novel acquisition protocols and the diversity of the data subjects collected from different geographic locations will allow developing a novel algorithm for either unimodal or multimodal biometrics.%
Further, we also report the performance evaluation of the baseline biometric verification and Presentation Attack Detection (PAD) on the newly collected dataset.
\end{abstract}


\IEEEpeerreviewmaketitle

\section{Introduction}
Secure and reliable access control using biometrics are deployed in various applications that include border control, smartphone unlocking, banking transactions, financial services, attendance system and etc.  The use of biometrics in access control applications not only improves the reliability but also improves the user experience by authenticating the user based on who they are. Thus, the user is neither required to remember passcode nor need to possess any smart cards to gain access to the control process. Based on ISO/IEC 2382-37, the concept of biometrics is defined as: ‘automated recognition of individuals based on their behavioural and biological characteristics \cite{ISO-IEC-2382-37-170206}’.  The biometric characteristics can be either a physical (face, iris, fingerprint, etc.)  or a behavioral (keystroke, gait, etc.) trait that can be used to recognize the data subject.

Evolution of the biometric technology has resulted in several consumer applications including smartphone biometrics. According to the Acuity Intelligence Forecast \cite{AcuityWebPage}, smartphone-based biometrics can surpass a revenue of 50.6 billion dollars by 2022 that also includes the revenue of biometric applications for financial transactions. Further, it is also estimated from TrendForce \cite{TrendWiseWebPage} that,  there will be 1.4 billion smartphone devices as of 2022. These factors indicate the evolution of different types of biometric-based applications for smartphones including banking applications in the lines of already available services from different vendors like Apple Pay, Samsung pay, Google Wallet.  The majority of commercial smartphones available today provide only  uni-modal biometrics and the most popular biometrics that are used by smartphone vendors include fingerprint, iris and face.

Even though the utility of biometrics on smartphones has enabled several advantages, there exist several challenges in real-life applications to take the full advantage of the biometrics authentication process on the consumer smartphone.  Among the many challenges, vulnerability towards attacks and interoperability challenges are key problems that limit reliable and secure applications of  smartphones for financial services. The vulnerability of smartphones to both direct (Presentation Attacks, aka.,  spoofing attacks) and indirect attacks are well exemplified in the literature \cite{Ramachandra:2017:PAD:3058791.3038924} \cite{marcel2018handbook}.  Further, it is also demonstrated in \cite{zhang2015fingerprints} that, the biometric template can be retrieved from the smartphone hardware chip, which further indicates another vulnerability. Recent works based on the master prints \cite{7893784} demonstrates the vulnerability of the fingerprint recognition system itself, irrespective of the manufacturer.  The second key challenge is the interoperability, as smartphone system uses proprietary biometric solutions, it is very challenging to use them with the traditional biometric systems. This limits the user by locking oneself to the particular smartphone to enable applications. These factors motivated the research towards smartphone biometrics that can be independent of the devices. To this extent, several attempts have been made to develop both uni-modal and multimodal biometric solution. The most popular biometric characteristics investigated include face \cite{RATTANI201839}, visible iris \cite{MICHEI}, soft-biometrics \cite{Patel}, finger photo \cite{FingerPhoto}.

The crucial aspect hindering the advances in the area of smartphone biometrics is the availability of suitable biometric data to benchmark the recognition performance of the newly developed algorithms and the reproducible evaluation. There exists a limited number of smartphone dataset which are publicly available and which are particularly addresssing biometric characteristics such as: face \cite{ 6266494}, fingerphoto \cite{FingerPhoto}, iris \cite{MICHEI}, soft-biometrics \cite{Patel} and multi-modality \cite{ bartuzi2018mobibits}. However, the collection of biometric data is a time and resource consuming tedious process that demands additional efforts in selecting the capture device, design of data collection protocols, post-processing of captured data, annotation of captured data, data anonymization and the collection itself. Further, one also needs to consider the legal regulations to obtain the data subject's consent and also to respect the data protection regulations.

This paper presents a recently collected smartphone based multimodal biometric dataset within the framework of the SWAN project\cite{SWANPage}. The SWAN Multimodal Biometric Dataset (SWAN-MBD) was collected between April 2016 till December 2017 with a collective effort from three different institutions: The Norwegian University of Science and Technology (NTNU), Norway, Idiap Research Institute, Switzerland and IDEMIA in France and India. The data collection is carried out in six different sessions with a time gap between sessions of 1 week to 3 weeks. The capture environment included both indoor and outdoor scenarios in assisted/supervised and unsupervised capture settings. The data capture includes both self and assisted capture processes replicating the real-life applications such as banking transactions. Three different biometric characteristics such as face, periocular and voice corresponding to 150 subjects were captured and the data subjects are living in 4 different geographic locations such as Norway, Switzerland, France and India.
Further, we also present a new SWAN Multimodal Presentation Attack Dataset (SWAN-MPAD) using two different types of Presentation Attack Instruments (PAIs) such as high-quality print and display attack (using iPhone 6 and iPad PRO) for the face and periocular characteristics. In the case of voice biometrics, two different quality loudspeakers were used to record the replay attacks on iPhone 6.  Being a new smartphone multimodal biometric dataset with presentation attack samples, it will allow one to develop and benchmark both verification and Presentation Attack Detection (PAD) algorithms.

The following are the main contribution of this paper:
\begin{itemize}
\item New multimodal dataset collected using smartphone in 6 different session from 150 subjects.
\item New multimodal presentation attack dataset collected using two different PAI on face, periocular, and voice biometrics.
\item Performance evaluation protocols to benchmark the accuracy of both verification and PAD algorithms.
\item  Quantitative result of the baseline algorithms are reported using ISO/IEC SC 37 metrics on both verification and PAD.
\end{itemize}

The rest of the paper is organised as follows: Section~\ref{sec:Rela} presents the related wok on the available multimodal biometric datasets collected using smartphone, Section \ref{sec:SWANdB} presents the SWAN multimodal biometric and presentation attack dataset, Section \ref{sec:Proto} presents the evaluation protocols to benchmark the algorithms, Section \ref{sec:baseline} discuss the different baseline systems used in benchmark the performance on SWAN multimodal biometric dataset. Section \ref{sec:Exp} discuss the quantitative results of the baseline systems, Section \ref{sec:PoT} discuss the potential use of newly collected dataset for research and development tasks, Section \ref{sec:DBDis} provides the information on data distribution and Section \ref{sec:Cocn} draws the conclusion.

\section{Related Work}
\label{sec:Rela}

\begin{table*}[htp]
  \centering
  \caption{Publicly available smartphone based multimodal datasets}
    \begin{tabular}{|c|c|p{8.07em}|c|c|c|}
    \hline
    \multicolumn{1}{|c|}{\multirow{2}[2]{*}{Dataset}} & \multicolumn{1}{c|}{\multirow{2}[2]{*}{Year}} & \multirow{2}[2]{*}{Devices} & \multicolumn{1}{c|}{\multirow{2}[2]{*}{No. of subjects}} & \multicolumn{1}{c|}{\multirow{2}[2]{*}{Biometric}} & \multicolumn{1}{c|}{\multirow{2}[2]{*}{Availability}} \bigstrut[t]\\
          &       & \multicolumn{1}{c|}{} &       &       &  \bigstrut[b]\\
    \hline
    \multicolumn{1}{|c|}{\multirow{2}[2]{*}{MOBIO \cite{6266494}}} & \multirow{2}[2]{*}{2012} & Nokia N93i & \multirow{2}[2]{*}{152} & \multicolumn{1}{c|}{\multirow{2}[2]{*}{Face, Voice}} & \multicolumn{1}{c|}{\multirow{2}[2]{*}{Free}} \bigstrut[t]\\
          &       & Mac-book &       &       &  \bigstrut[b]\\
    \hline
    \multicolumn{1}{|c|}{\multirow{2}[2]{*}{CSIP \cite{CSIAP_DB}}} & \multirow{2}[2]{*}{2015} & Sony Xperia & \multirow{2}[2]{*}{50} & \multicolumn{1}{c|}{\multirow{2}[2]{*}{Iris, Periocular}} & \multicolumn{1}{c|}{\multirow{2}[2]{*}{Free}} \bigstrut[t]\\
          &       & Apple IPhone 4 &       &       &  \bigstrut[b]\\
    \hline
    \multicolumn{1}{|c|}{\multirow{2}[2]{*}{FTV \cite{FTV_DB}}} &  \multirow{2}[2]{*}{2010} & \multirow{2}[2]{*}{HP iPAQ rw6100} & \multirow{2}[2]{*}{50} & \multicolumn{1}{c|}{\multirow{2}[2]{*}{Face, Teeth, Voice}} & \multicolumn{1}{c|}{\multirow{2}[2]{*}{Free}} \bigstrut[b]\\
     &       &    &       &       &  \bigstrut[b]\\
    \hline
     \multicolumn{1}{|c|}{\multirow{2}[2]{*}{MobBIO \cite{MobBIO}}} &  \multirow{2}[2]{*}{2014} & \multirow{2}[2]{*}{ASUS PAD TF 300} & \multirow{2}[2]{*}{105} & \multicolumn{1}{c|}{\multirow{2}[2]{*}{Voice, Face, periocular}} & \multicolumn{1}{c|}{\multirow{2}[2]{*}{Free}} \bigstrut[b]\\
&       &    &       &       &  \bigstrut[b]\\
  \hline
 \multicolumn{1}{|c|}{\multirow{2}[2]{*}{UMDAA \cite{UMDAA}}} &  \multirow{2}[2]{*}{2016} & \multirow{2}[2]{*}{Nexus 5} & \multirow{2}[2]{*}{48} & \multicolumn{1}{c|}{\multirow{2}[2]{*}{Face, Behavior patterns}} & \multicolumn{1}{c|}{\multirow{2}[2]{*}{Free}} \bigstrut[b]\\
&       &    &       &       &  \bigstrut[b]\\
    \hline
    \multicolumn{1}{|c|}{\multirow{3}[2]{*}{MobiBits \cite{MobiBits_DB}}} & \multirow{3}[2]{*}{2018} & Huawei Mate S & \multirow{3}[2]{*}{53} & \multicolumn{1}{c|}{\multirow{3}[2]{*}{Signature, Voice, Face, Periocular, Hand}} & \multicolumn{1}{c|}{\multirow{3}[2]{*}{Free}} \bigstrut[t]\\
          &       & Huawei P9 Lite &       &       &  \\
          &       & CAT S60 &       &       &  \bigstrut[b]\\
    \hline
    \multicolumn{1}{|c|}{\multirow{5}[2]{*}{BioSecure-DS3 \cite{BioSecure_DB}}} & \multirow{5}[2]{*}{2010} & Samsung Q1 & \multirow{5}[2]{*}{713} & \multicolumn{1}{c|}{\multirow{5}[2]{*}{Voice, Signature, Face, Fingerprint}} & \multicolumn{1}{c|}{\multirow{5}[2]{*}{Paid}} \bigstrut[t]\\
          &       & Philips SP900NC &       &       &  \\
          &       & Webcam &       &       &  \\
          &       & HP iPAQ hx2790 &       &       &  \\
          &       & PDA   &       &       &  \bigstrut[b]\\
    \hline
    \hline
 \multicolumn{1}{|c|}{\multirow{2}[2]{*}{SWAN Dataset}} &  \multirow{2}[2]{*}{2019} & \multirow{2}[2]{*}{iPhone 6S} & \multirow{2}[2]{*}{150} & \multicolumn{1}{c|}{\multirow{2}[2]{*}{Face,  Periocular, Multilingual Voice}} & \multicolumn{1}{c|}{\multirow{2}[2]{*}{Free}} \\
 &       &    &       &        &  \bigstrut[t]\\
&       &    &       &    Presentation Attack dataset    &  \bigstrut[b]\\
 \hline
    \end{tabular}%
  \label{tab:TableDB}%
\end{table*}%

In this section, we discuss the publicly available smartphone based multimodal datasets. There are limited smartphones based multimodal biometric databases currently available for researchers.  The majority of these available datasets are based on the two modalities that can be captured together, for example, face (talking) and voice, iris and periocular. Table \ref{tab:TableDB}  presents an overview of the different smartphone based multimodal datasets that are publicly available.

The \textbf{BioSecure} dataset (DS-3) \cite{BioSecure_DB} is one of the earlier and large scale publicly available datasets (available upon license fee payment).   The BioSecure dataset is comprised of four different faces, voice, signature and fingerprint collected from 713 subjects in 2 sessions. Only face and voice are collected using the mobile device Samsung Q1.

The \textbf{MOBIO} database \cite{6266494} is comprised of 153 subjects from which the biometric characteristics  (face and voice) are collected using Nokia N93i  and MacBook laptop. The complete dataset is collected in 12 sessions by capturing the face (talking) together with voice. Voice samples are collected based on both pre-defined text and free text, and the face biometrics is recorded while the subject is talking.

The \textbf{CSIP} database \cite{CSIAP_DB} is comprising of 50 subjects with iris and periocular biometric characteristics captured using four different smartphone devices. The entire dataset is collected with different backgrounds to reflect the real-life scenario.

The \textbf{FTV} dataset \cite{FTV_DB} consists of face, teeth and voice biometric characteristics captured from 50 data subjects using a smartphone HP iPAQ rw6100. The face and teeth samples are collected using the smartphone camera while the microphone of smartphone used to collect the voice samples.

The \textbf{MobBIO} dataset \cite{MobBIO} is based on the three different biometric characteristics such as the face, voice and periocular using a tablet Asus Transformer Pad TF 300T. Voice samples are collected using a microphone of Asus Transformer Pad TF 300T in which data subject was asked to readout 16 sentences in Portuguese. The face and periocular samples are collected using the 8MP camera from Asus Transformer Pad TF 300T. This dataset is comprised of 105 data subjects collected in two different lighting conditions.

The \textbf{UMDAA} dataset \cite{UMDAA} is collected using the Nexus 5 from 48 data subjects. This database has a collection of both physical and behavioral biometric characteristics.  The data collection sensors include the front-facing camera, touchscreen, gyroscope, magnetometer, light sensor, GPS, Bluetooth, accelerometer, WiFi, proximity sensor, temperature sensor and pressure sensor. The entire dataset is collected for two months and provides face and other related behavior patterns suitable for continuous authentication.

The \textbf{MobiBits}  dataset \cite{MobiBits_DB} consists of five different biometric characteristics namely voice, face, iris, hand and signature that are collected from 55 data subjects. Three different smartphones are used to collect the biometric characteristics such as Huawei Mate smartphone is used to collect signature and periocular samples, CAT S60 smartphone is used to collect hand and face samples, Huawei P9 Lite smartphone is used to collect voice samples.

\subsection{Features of the SWAN Multimodal Biometric Dataset}
\label{Sec:DBUnique}
SWAN Multimodal Biometric Dataset (SWAN-MBD) is collected to complement the existing datasets and the data collection protocols are designed to meet the real-life scenario such as banking transactions. The following are the main features of the newly introduced SWAN Multimodal Biometric Dataset:
\begin{itemize}
\item The data collection protocol is designed to capture data in 6 different sessions. Each session reflects both time variation and capturing conditions (outdoor and indoor).
\item The data collection application is developed to make  data collection consistent with ease of use in regards to the user interaction such that self-capture is facilitated.  This is the unique feature of this dataset in which the data is self-captured by the participants.
\item The data collection is carried out in four different geographic locations such as India, Norway, France and Switzerland with multiple ethnicity representation.
\item Three different biometric characteristics: face, periocular and voice are captured from 150 data subjects in all 6 different sessions. The voice samples are captured based on the question and answers which is text-dependent. The voice samples are collected in four different languages: English, Norwegian, French and Hindi.
\item The whole dataset is collected using iPhone 6 smartphone and iPad-PRO.
\item In addition, we also present a new SWAN Multimodal Presentation Attack dataset (SWAN-MPAD) for all three biometric characteristics (face, periocular and voice).
\end{itemize}

\section{SWAN Multimodal Biometric Dataset}
\label{sec:SWANdB}
\subsection{Database Acquisition}
\begin{figure}[htp]
\centering
 	\includegraphics[width= 1.0\linewidth]{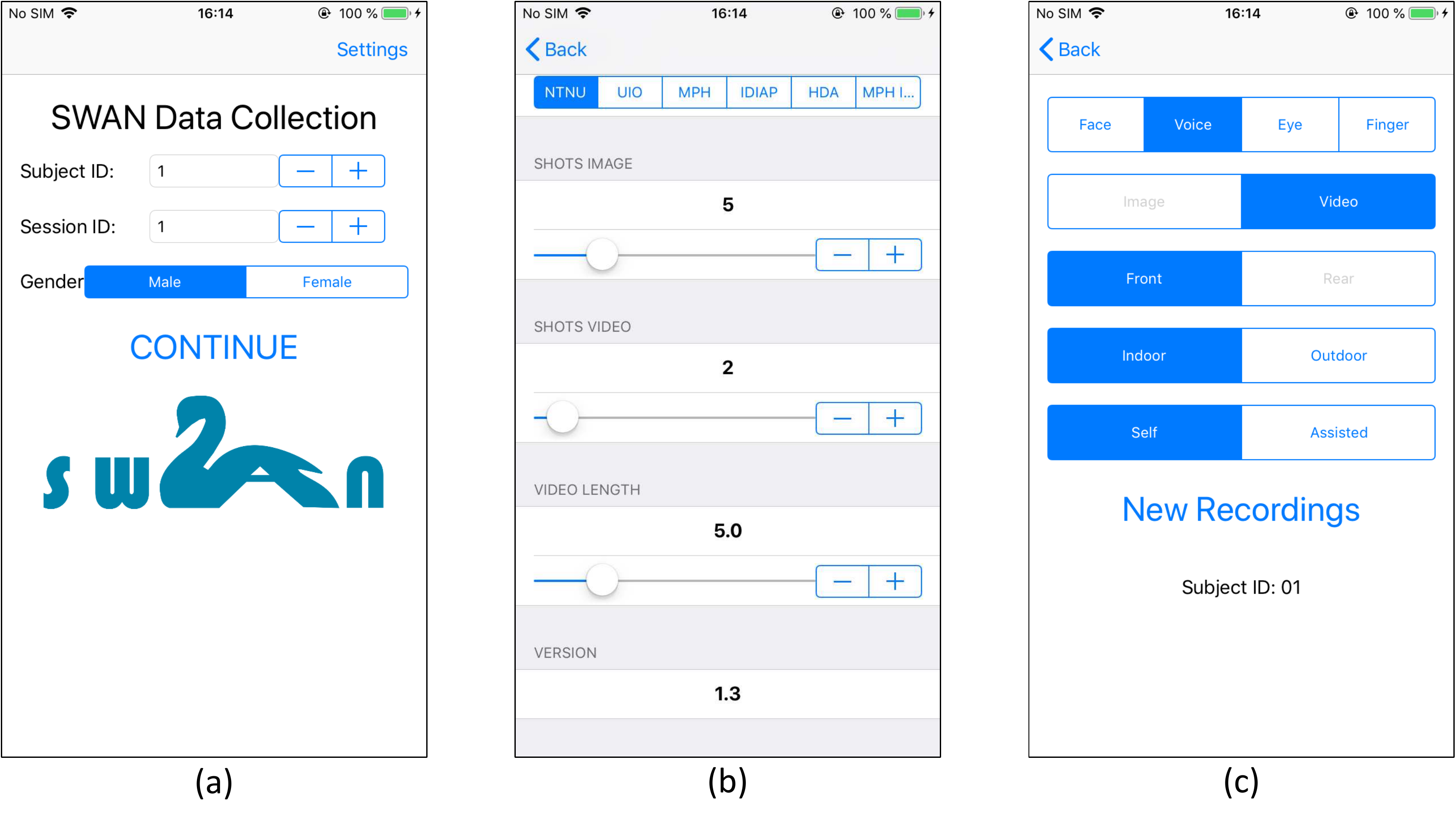}
 	\caption{Screen shot from SWAN multimodal biometric data capture application}
 	\label{fig:DatcaptureGUI}
\end{figure}

\begin{figure}[htp]
\centering
 	\includegraphics[width= 1.0\linewidth]{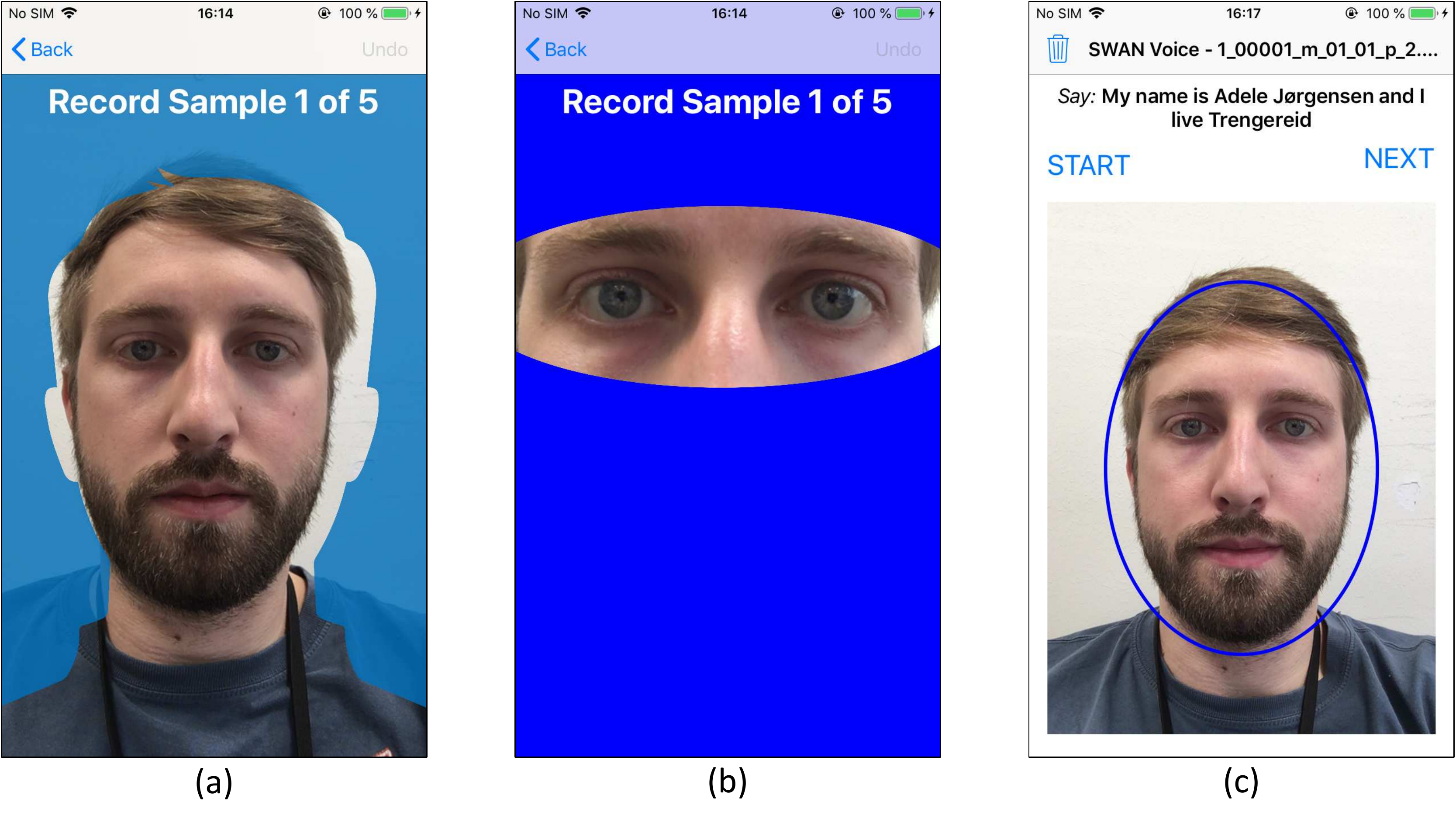}
 	\caption{Illustration of SWAN multimodal biometric data capture application during (a) Face capture (b) Eye region capture (c) Voice and talking face capture}
 	\label{fig:DatcaptureBio}
\end{figure}

To facilitate the data collection at different locations, a smartphone application is developed for the iOS platform (version 9) that can be installed in the data capture devices (both iPhone and iPad Pro). This application has a Graphical User Interface that allows the data collection moderator to select session number, biometric characteristics, location ID, subject ID and other relevant information for data collection. Figure \ref{fig:DatcaptureGUI} shows the GUI of SWAN data collection application while Figure \ref{fig:DatcaptureBio} shows the example images during biometric data collection. Thus, the application is designed to make sure the data collection process can be easily carried out such that data subjects can use it seamlessly during the self-capture protocol.  The data collection process is broadly divided into two phases as explained below. %

\subsection{SWAN Multimodal Biometric Dataset}
The SWAN multimodal biometric dataset was jointly collected within SWAN project by  partners at four different geographic locations: Norway, Switzerland, France and India. The dataset is  comprised of 150 subjects such that 50 data subjects collected in Norway, 50 data subjects are collected in Switzerland, 5 data subjects in France and 45 data subjects in India.  The age distribution of the Data subjects are between 20 to 60 years. The dataset was collected using Apple iPhone6S and Apple iPad Pro (12.9‑inch iPad Pro). Three biometric modalities are collected: (1) Face: both Ultra HD images and HD (including slow motion) video recordings of faces from data subjects (2) Voice: HD audio-visual recordings of talking faces from data subjects (3) Eye:  both Ultra HD images and HD (including slow motion) video recordings of eyes from data subjects.  The whole dataset is collected in 6 different sessions such that, \textbf{Session 1} is captured in an indoor environment with uniform illumination of the face, quiet environment reflecting high quality supervised enrolment. \textbf{Session 2} is captured in indoor with natural illumination on the face and semi-quite environment. \textbf{Session 3} is captured in outdoor with uncontrolled illumination, natural noise environment. \textbf{Session 4} is captured in indoor with uncontrolled illumination e.g., side light from windows, natural noise environment. \textbf{Session 5} is captured in indoor with uncontrolled illumination, natural noise environment in crowded place. \textbf{Session 6} is captured in indoor with uncontrolled illumination e.g., side light from windows, natural noise environment. The average time duration between sessions varies from 1 week to 3 weeks.
The multimodal biometric samples are captured in both assisted capture mode (where data subjects are assisted during capturing) and self-capture mode (here data subject control the capture process on their own). Multimodal biometrics are collected using both rear and front cameras from iPhone and iPad Pro.
The iPhone 6S is used as the primary device to collect the dataset, while the iPad Pro is used to collect the data only in session-1 to capture a high-quality image that can be used to generate a Presentation Attack Instrument.

\begin{table}[htbp]
  \centering
  \caption{SWAN multimodal dataset: Face biometrics data collection}
  \resizebox{\columnwidth}{!}{
    \begin{tabular}{|c|p{3em}|p{6em}|p{6em}|p{15.335em}|}
    \hline
    \multicolumn{1}{|p{5em}|}{Modality} & Session & Capture Mode & Device /camera & Data capture per data subject \bigstrut\\
    \hline
    \multicolumn{1}{|c|}{\multirow{14}[18]{*}{Face}} & \multirow{9}[8]{*}{S1} & \multirow{5}[4]{*}{Assisted} & \multirow{3}[2]{*}{iPhone6S/Rear} & 5 Images (4032x3024, PNG) \bigstrut[t]\\
          & \multicolumn{1}{c|}{} & \multicolumn{1}{c|}{} & \multicolumn{1}{c|}{} & 2 videos (1280x720, 240fps, 5s, MP4)   \\
          & \multicolumn{1}{c|}{} & \multicolumn{1}{c|}{} & \multicolumn{1}{c|}{} & \multicolumn{1}{c|}{} \bigstrut[b]\\
\cline{4-5}          & \multicolumn{1}{c|}{} & \multicolumn{1}{c|}{} & \multirow{2}[2]{*}{iPad Pro/ Rear} & 5 Images (4032x3024, PNG) \bigstrut[t]\\
          & \multicolumn{1}{c|}{} & \multicolumn{1}{c|}{} & \multicolumn{1}{c|}{} & 2 videos (1280x720, 120fps, 5s, MP4)   \bigstrut[b]\\
\cline{3-5}          & \multicolumn{1}{c|}{} & \multirow{4}[4]{*}{Self-Capture} & \multirow{2}[2]{*}{iPhone6S/Front} & 5 Images (2576x1932, PNG) \bigstrut[t]\\
          & \multicolumn{1}{c|}{} & \multicolumn{1}{c|}{} & \multicolumn{1}{c|}{} & 2 videos (1280x720, 30fps, 5s, MP4) \bigstrut[b]\\
\cline{4-5}          & \multicolumn{1}{c|}{} & \multicolumn{1}{c|}{} & \multirow{2}[2]{*}{iPad Pro/ Front} & 5 Images (2576x1932, PNG) \bigstrut[t]\\
          & \multicolumn{1}{c|}{} & \multicolumn{1}{c|}{} & \multicolumn{1}{c|}{} & 2 videos (1280x720, 30fps, 5s, MP4) \bigstrut[b]\\
\cline{2-5}          & S2    & Self-Capture  & iPhone6S/Front & 2 videos (1280x720, 30fps, 5s, MP4) \bigstrut\\
\cline{2-5}          & S3    & Self-Capture  & iPhone6S/Front & 2 videos (1280x720, 30fps, 5s, MP4) \bigstrut\\
\cline{2-5}          & S4    & Self-Capture  & iPhone6S/Front & 2 videos (1280x720, 30fps, 5s, MP4) \bigstrut\\
\cline{2-5}          & S5    & Self-Capture  & iPhone6S/Front & 2 videos (1280x720, 30fps, 5s, MP4) \bigstrut\\
\cline{2-5}          & S6    & Self-Capture  & iPhone6S/Front & 2 videos (1280x720, 30fps, 5s, MP4) \bigstrut\\
    \hline
    \end{tabular}%
  \label{tab:SWANDB_Face}%
      }
\end{table}%

\begin{table}[htbp]
  \centering
  \caption{SWAN multimodal dataset: Eye region biometrics data collection}
  	\resizebox{\columnwidth}{!}{
    \begin{tabular}{|c|c|c|c|p{15.915em}|}
    \hline
    \multicolumn{1}{|p{6.335em}|}{Modality} & \multicolumn{1}{p{5em}|}{Session} & \multicolumn{1}{p{6em}|}{Capture Mode} & \multicolumn{1}{p{6.585em}|}{Device /camera} & Data capture per data subject \bigstrut\\
    \hline
    \multicolumn{1}{|c|}{\multirow{29}[28]{*}{Eye}} & \multicolumn{1}{c|}{\multirow{9}[8]{*}{S1}} & \multicolumn{1}{c|}{\multirow{5}[4]{*}{Assisted}} & \multicolumn{1}{c|}{\multirow{3}[2]{*}{iPhone6S/Rear}} & 5 Images (4032x3024, PNG) \bigstrut[t]\\
          &       &       &       & \multirow{2}[1]{*}{2 videos (1280x720, 240fps, 5s, MP4)} \\
          &       &       &       & \multicolumn{1}{c|}{} \bigstrut[b]\\
\cline{4-5}          &       &       & \multicolumn{1}{c|}{\multirow{2}[2]{*}{iPad Pro/Rear}} & 5 Images (4032x3024, PNG) \bigstrut[t]\\
          &       &       &       & 2 videos (1280x720, 120fps, 5s, MP4) \bigstrut[b]\\
\cline{3-5}          &       & \multicolumn{1}{c|}{\multirow{4}[4]{*}{Self-Capture}} & \multicolumn{1}{c|}{\multirow{2}[2]{*}{iPhone6S/Front}} & 5 Images (2576x1932, PNG) \bigstrut[t]\\
          &       &       &       & 2 videos (1280x720, 30fps, 5s, MP4) \bigstrut[b]\\
\cline{4-5}          &       &       & \multicolumn{1}{c|}{\multirow{2}[2]{*}{iPad Pro/Front}} & 5 Images (2576x1932, PNG) \bigstrut[t]\\
          &       &       &       & 2 videos (1280x720, 30fps, 5s, MP4) \bigstrut[b]\\
\cline{2-5}          & \multicolumn{1}{c|}{\multirow{4}[4]{*}{S2}} & \multicolumn{1}{c|}{\multirow{2}[2]{*}{Assisted}} & \multicolumn{1}{c|}{\multirow{2}[2]{*}{iPhone6S/Rear}} & 5 Images (4032x3024, PNG) \bigstrut[t]\\
          &       &       &       & 2 videos (1280x720, 240fps, 5s, MP4) \bigstrut[b]\\
\cline{3-5}          &       & \multicolumn{1}{c|}{\multirow{2}[2]{*}{Self-Capture}} & \multicolumn{1}{c|}{\multirow{2}[2]{*}{iPhone6S/Front}} & 5 Images (2576x1932, PNG) \bigstrut[t]\\
          &       &       &       & 2 videos (1280x720, 30fps, 5s, MP4) \bigstrut[b]\\
\cline{2-5}          & \multicolumn{1}{c|}{\multirow{4}[4]{*}{S3}} & \multicolumn{1}{c|}{\multirow{2}[2]{*}{Assisted}} & \multicolumn{1}{c|}{\multirow{2}[2]{*}{iPhone6S/Rear}} & 5 Images (4032x3024, PNG) \bigstrut[t]\\
          &       &       &       & 2 videos (1280x720, 240fps, 5s, MP4) \bigstrut[b]\\
\cline{3-5}          &       & \multicolumn{1}{c|}{\multirow{2}[2]{*}{Self-Capture}} & \multicolumn{1}{c|}{\multirow{2}[2]{*}{iPhone6S/Front}} & 5 Images (2576x1932, PNG) \bigstrut[t]\\
          &       &       &       & 2 videos (1280x720, 30fps, 5s, MP4) \bigstrut[b]\\
\cline{2-5}          & \multicolumn{1}{c|}{\multirow{4}[4]{*}{S4}} & \multicolumn{1}{c|}{\multirow{2}[2]{*}{Assisted }} & \multicolumn{1}{c|}{\multirow{2}[2]{*}{iPhone6S/Rear}} & 5 Images (4032x3024, PNG) \bigstrut[t]\\
          &       &       &       & 2 videos (1280x720, 240fps, 5s, MP4) \bigstrut[b]\\
\cline{3-5}          &       & \multicolumn{1}{c|}{\multirow{2}[2]{*}{Self-Capture}} & \multicolumn{1}{c|}{\multirow{2}[2]{*}{iPhone6S/Front}} & 5 Images (2576x1932, PNG) \bigstrut[t]\\
          &       &       &       & 2 videos (1280x720, 30fps, 5s, MP4) \bigstrut[b]\\
\cline{2-5}          & \multicolumn{1}{c|}{\multirow{4}[4]{*}{S5}} & \multicolumn{1}{c|}{\multirow{2}[2]{*}{Assisted}} & \multicolumn{1}{c|}{\multirow{2}[2]{*}{iPhone6S/Rear}} & 5 Images (4032x3024, PNG) \bigstrut[t]\\
          &       &       &       & 2 videos (1280x720, 240fps, 5s, MP4) \bigstrut[b]\\
\cline{3-5}          &       & \multicolumn{1}{c|}{\multirow{2}[2]{*}{Self-Capture}} & \multicolumn{1}{c|}{\multirow{2}[2]{*}{iPhone6S/Front}} & 5 Images (2576x1932, PNG) \bigstrut[t]\\
          &       &       &       & 2 videos (1280x720, 30fps, 5s, MP4) \bigstrut[b]\\
\cline{2-5}          & \multicolumn{1}{c|}{\multirow{4}[4]{*}{S6}} & \multicolumn{1}{c|}{\multirow{2}[2]{*}{Assisted}} & \multicolumn{1}{c|}{\multirow{2}[2]{*}{iPhone6S/Rear}} & 5 Images (4032x3024, PNG) \bigstrut[t]\\
          &       &       &       & 2 videos (1280x720, 240fps, 5s, MP4) \bigstrut[b]\\
\cline{3-5}          &       & \multicolumn{1}{c|}{\multirow{2}[2]{*}{Self-Capture}} & \multicolumn{1}{c|}{\multirow{2}[2]{*}{iPhone6S/Front}} & 5 Images (2576x1932, PNG) \bigstrut[t]\\
          &       &       &       & 2 videos (1280x720, 30fps, 5s, MP4) \bigstrut[b]\\
    \hline
    \end{tabular}%
  \label{tab:SWANDB_Eye}%
  }
\end{table}%

\begin{table}[htbp]
  \centering
  \caption{SWAN multimodal dataset: Voice and Talking face biometrics data collection}
\resizebox{\columnwidth}{!}{%
\begin{tabular}{|l|l|l|l|l|}
\hline
Modality                                                          & Session & Capture Mode & Device/Camera  & Data capture per data subject                                                                                                                     \\ \hline
\begin{tabular}[c]{@{}l@{}}Voice and\\ Talking faces\end{tabular} & s1-6    & Self-capture & iPhone6S/Front & \begin{tabular}[c]{@{}l@{}}8 videos: 4 videos in English\\ and 4 videos in native language\\ (1280x720, 30fps, variable length, MP4)\end{tabular} \\ \hline
\end{tabular}%
\label{tab:SWANDB_Voice}%
  }
\end{table}%
\begin{table*}[htp]
\centering
\caption{SWAN Presentation attack data collection details}
\label{tab:pai}
\resizebox{\textwidth}{!}{%
\begin{tabular}{|l|l|l|l|l|l|l|}
\hline
\multirow{2}{*}{Modality} & \multirow{2}{*}{PA name} & \multicolumn{3}{l|}{PA source selection} & \multirow{2}{*}{PAI}                                                                                           & \multirow{2}{*}{Biometric capture sensor} \\ \cline{3-5}
                          &                          & Session & Sensor                & type   &                                                                                                                &                                           \\ \hline
\multirow{3}{*}{Face}     & PA.F.1                   & 1          & { iPhone 6s back camera}    & photos                        & \begin{tabular}[c]{@{}l@{}}Epson Expression Photo XP-860\\ Epson Photo Paper Glossy, PN: S041271\end{tabular}  & iPhone 6s front camera (video)            \\ \cline{2-7}
                          & PA.F.5                   & 1          & { iPhone 6s front camera}    & videos                        & iPhone 6s display                                                                                              & iPhone 6s front camera (video)            \\ \cline{2-7}
                          & PA.F.6                   & 1          & { iPad-Pro front camera}     & videos                        & iPad-Pro display                                                                                               & iPhone 6s front camera (video)            \\ \hline
\multirow{3}{*}{Eye}      & PA.EI.1                  & 1          & { iPhone 6s back camera}    & photos                        & \begin{tabular}[c]{@{}l@{}}Epson Expression Photo XP-860\\ Epson Photo Paper Glossy, PN:  S041271\end{tabular} & iPhone 6s front camera (video)            \\ \cline{2-7}
                          & PA.EI.4                  & 1          & { iPad-Pro front camera}     & photos                        & iPad-Pro display                                                                                               & iPhone 6s front camera (video)            \\ \cline{2-7}
                          & PA.EI.5                  & 1          & { iPhone 6s front camera}    & videos                        & iPhone 6s display                                                                                              & iPhone 6s front camera (video)            \\ \hline
\multirow{2}{*}{Voice}    & PA.V.4                   & 1          & iPad-Pro microphone     & audio                         & Logitech high quality loudspeaker                                                                   & iPhone 6s microphone (audio)              \\ \cline{2-7}
                          & PA.V.7                   & 1          & iPhone 6s microphone     & audio                         & iPhone 6s speakers                                                                                              & iPhone 6s microphone (audio)              \\ \hline
\end{tabular}%
  \label{tab:SWANDB_PA}%
}
\end{table*}

Table \ref{tab:SWANDB_Face} indicates the data collection protocol and the sample collection (both images and videos) for the facial biometric characteristic. The biometric face capture indicated in Table \ref{tab:SWANDB_Face}  corresponds to one data subject. iPhone 6S is used to capture the face data in all six session and iPad Pro is used only in session 1.  During each acquisition of session-1, the face data is captured in both assisted and self-captured mode using rear and front camera from iPhone6S and iPad Pro respectively. The data collected using the iPad Pro is used to generate presentation attacks while data collected from iPhone6S is used to perform the biometric verification. Thus, in total there are: $150\times$ Subjects $\times20$ images = $3000$ image and  $150\times$ Subjects $\times18$ videos = $2700$ videos.  Table \ref{tab:SWANDB_Eye} presents the data statistics corresponding to eye region biometric characteristic, which is in both self-capture and assisted mode. Since the goal of the eye region data collection is to get high quality images that can be used for both periocular and visible iris recognition, we collected the eye region dataset in both capture modes across all 6 sessions. Therefore the assisted mode is captured using the rear camera of iPhone 6S with 12mega pixels that provides good quality images for visible iris recognition. However, the images collected from the self-capture process using the frontal camera can be used to develop the periocular verification systems. Similar to face capture process, the iPad Pro is used only in session 1 to capture a good quality eye region image that is used to generate a presentation attack instrument, to be used against the periocular biometric system. The whole dataset consists of $150 \times $ Subjects $\times70$ images = $10500$ image and $150\times $ Subjects $\times28$ videos = $4200$ videos.

\begin{figure}[htp]
\centering
 	\includegraphics[width= 1\linewidth]{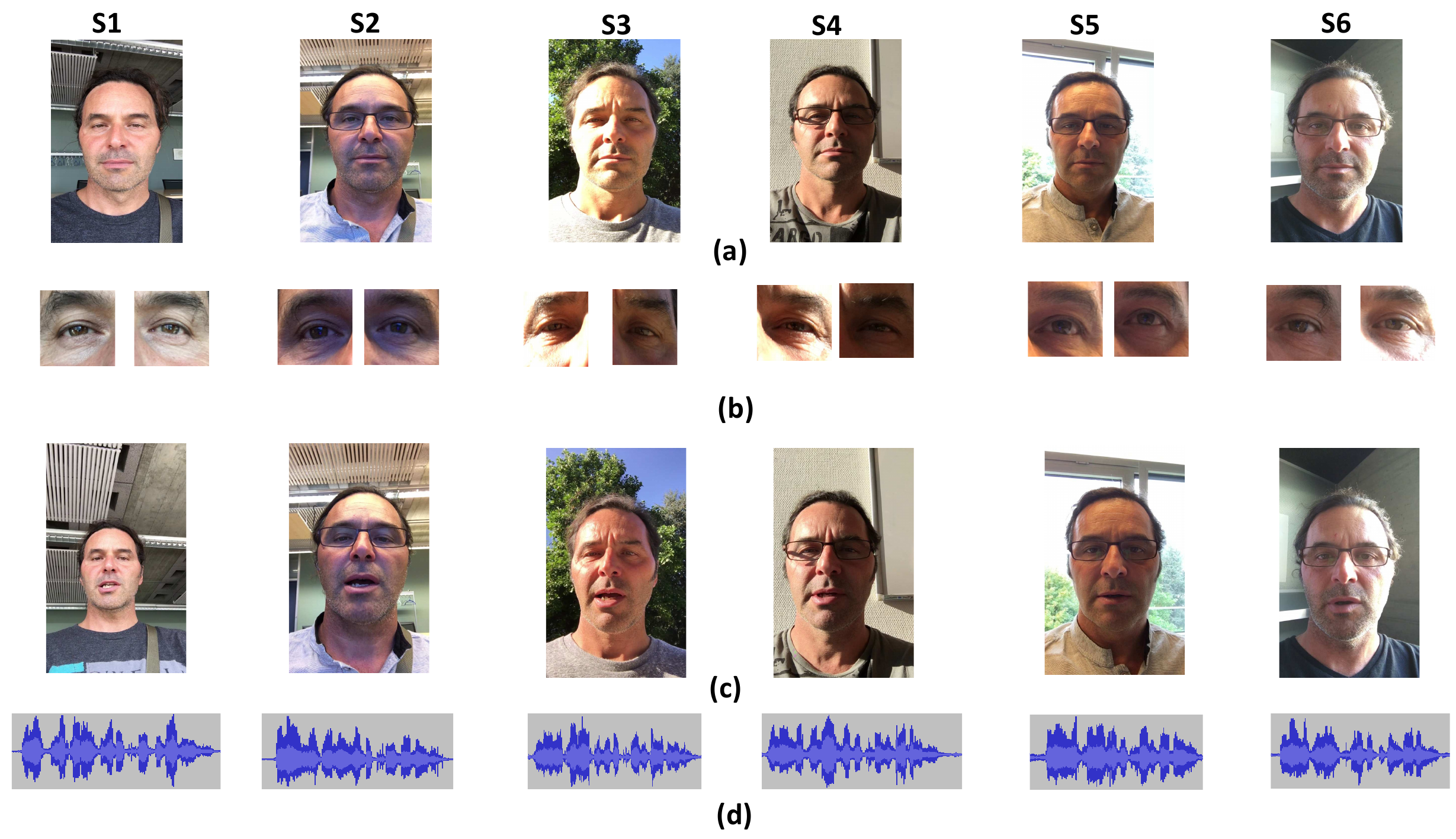}
 	\caption{Illustration of SWAN multimodal biometric dataset images (a) Face capture (b) Eye region capture (c) Talking face capture (one only frame is included for the simplicity) (d) voice sample}
 	\label{fig:databaseImagesAll}
\end{figure}

Table \ref{tab:SWANDB_Voice} indicates the statistics and protocol of the voice and talking face data collection. To cover both text independent and text dependent modes, the audio recordings (actually audio-video recordings) are captured with the data subjects pronouncing these 4 utterances in English followed by these 4 utterances in a national language depending on the site (Norwegian, French and Hindi).  The four sentences include: \textit{Sentence \#1}: "My name is \texttt{FAKE\_FIRSTNAME} \texttt{FAKE\_NAME} and I live \texttt{FAKE\_ADDRESS}" the address is a short string that is limited to the street name. Thus no street number, no zip code, no city and no country information is recorded. \textit{Sentence \#2}: "My account number is \texttt{FAKE\_ACCOUNTNUMBER}". The account numbers are presented by groups of digits (eg. "5354" "8745"). The data subject is free to pronounce the groups of digits the way he/she wants (digits by digits, as a single number or as a combination).  \textit{Sentence \#3:} "The limit of my card is $5000$ euros". \textit{Sentence \#4:}  "The code is $9$ $8$ $7$ $6$ $5$ $4$ $3$ $2$ $1$ $0$".  The $10$ digits are presented one by one and the data subjects were asked to pronounce the digits separately. The Audio-visual data with voice and talking faces are collected in the self-capture mode using the frontal camera of iPhone6S in all 6 sessions. Thus, $150\times$ Subjects $\times48$ videos = $7200$ videos corresponding to audio-visual data.  Figure \ref{fig:databaseImagesAll} illustrates the example images from SWAN multimodal dataset collected in all six session indicating an inter-session variability.

\subsection{SWAN-Presentation Attack Dataset}
The SWAN presentation attack dataset is comprised of three different types of presentation attacks that are generated for three different biometric characteristics, namely: face, eye and voice. The presentation attack (PA) database generation generally requires obtaining
biometric samples "PA source selection", generating artefacts, and presenting attack artefacts to the biometric capture sensor.  Session 1 recordings from SWAN multibiometric dataset is used to generate the presentation attack dataset.  The artefact generation is carried out using five different Presentation Attack Instruments (PAI) such as: (1)High quality photo on paper generated using the photo printer (face \& Eye). The print artefacts on face and eye are generated using Epson Expression Photo XP-860 with high quality paper Epson Photo Paper Glossy ({Product Number:}  S041271; {Basis Weight:} 52 lb.($200 g/m^2$); {Size:} A4; {Thickness:} 8.1 mil.). (2)  Electronic display of artefacts using iPad-Pro (face \& Eye)  (3) Electronic display of artefacts using iPhone 6S (face \& Eye) (4) Logitech high quality loudspeaker (Voice) (5) iPad-Pro loudspeaker (Voice).  We have used these PAI by considering the cost for generation versus attack potential and thus selected PAIs, which are of low-cost (reasonably) and at the same time indicate a high vulnerability for the biometric system under attack.

\begin{figure}[htb]
\centering
 	\includegraphics[width= 1\linewidth]{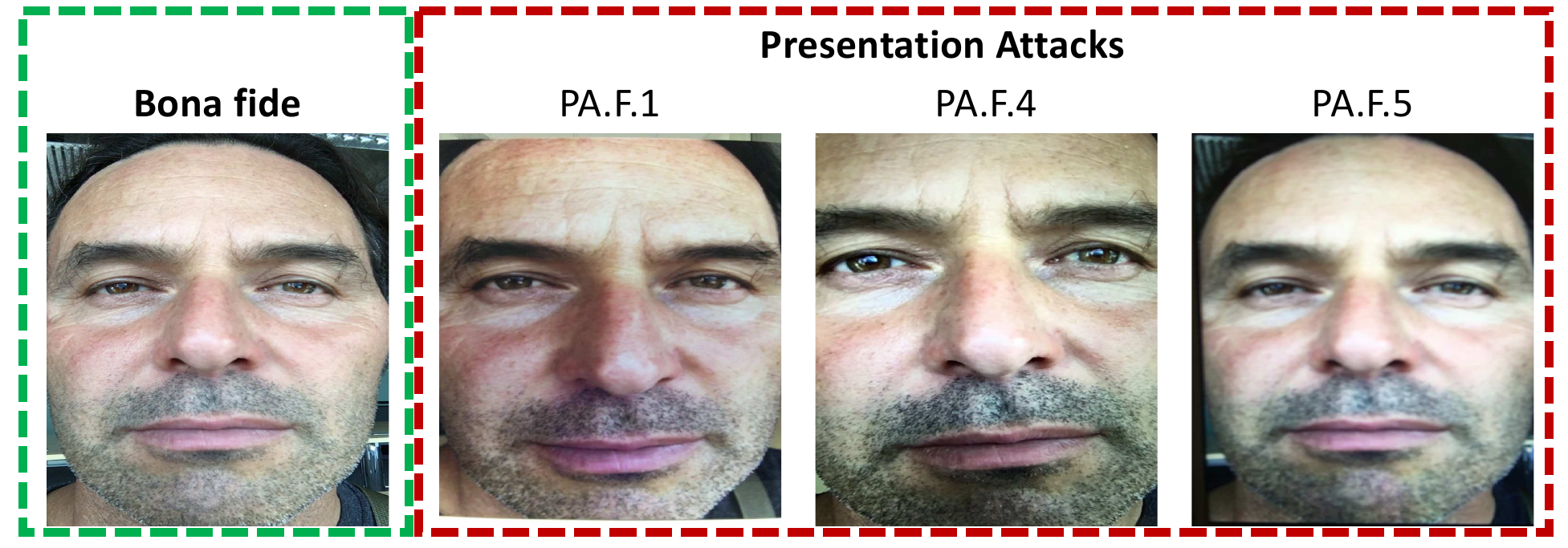}
 	\caption{Illustration of SWAN Presentation Attack dataset  (a) Bona fide (b) Presentation Attack (PA.F.1) (c) Presentation Attack (PA.F.4) (d) Presentation Attack (PA.F.5)}
 	\label{fig:PADFaceImagesAll}
\end{figure}

\begin{figure}[htb]
\centering
 	\includegraphics[width= 1\linewidth]{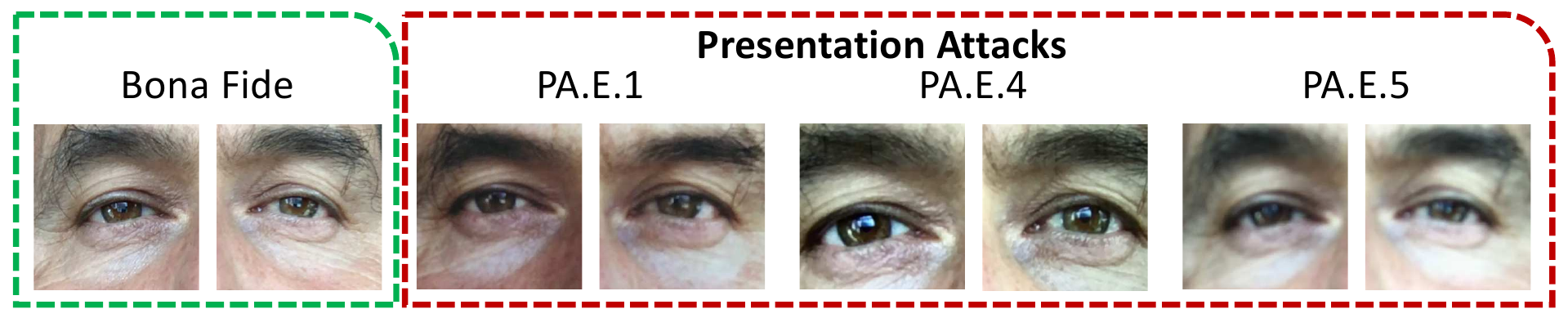}
 	\caption{Illustration of SWAN Presentation Attack dataset  (a) Bona fide (b) Presentation Attack (PA.E.1) (c) Presentation Attack (PA.E.4) (d) Presentation Attack (PA.E.5)}
 	\label{fig:PADEyeImagesAll}
\end{figure}

Table \ref{tab:SWANDB_PA} presents the PAD data collection procedure with presentation attack source, PAI and biometric capture sensor. All the presentation attack data for face and eye are collected using a frontal camera of iPhone 6S and all recordings were videos and at least 5 seconds long for each PA and in the \emph{.mp4} format. Both the biometric capture device (iPhone 6s) and the PAI (paper or display) is mounted on stands.  In the case of voice samples, the audio files collected using iPhone6S with the same compression specifications as bona fide samples. Figure \ref{fig:PADFaceImagesAll} \& Figure \ref{fig:PADEyeImagesAll} shows the example images for face and eye presentation attacks respectively.

\section{Experimental Performance Evaluation Protocols}
\label{sec:Proto}
In this section, we discuss the evaluation protocol that is used to report the performance of the baseline algorithm for both verification and Presentation Attack Detection (PAD) subsystems.
\subsection{Biometric verification performance evaluation protocol}
To evaluate the biometric verification performance, we propose two protocols: \textbf{Protocol-1:} designed to study the individual verification performance from independent sessions. Thus, we enrol images from session 2 as reference samples and then compute the verification performance using samples from session 3 to session 6 individually as probe samples. \textbf{Protocol-2:} this protocol is designed to evaluate the performance of the biometric system across all session.  Thus, in this protocol, we enrol images from session-2  and probe biometric samples from all session 3 to session 6, to evaluate the verification performance. However, for simplicity, we have used only self-captured data to report the performance of the baseline verification systems.
\subsection{Presentation Attack Detection protocol}
To evaluate the performance of the baseline presentation attack detection techniques we propose two different evaluation protocols. \textbf{Protocol-1}: This protocol is to evaluate the performance of the PAD techniques independently on each PAI, thus training and testing of the PAD techniques are carried out independently on each PAI. \textbf{Protocol-2}: This protocol is designed to evaluate the performance of the PAD algorithms when trained and tested with all PAIs. To effectively evaluate the PAD algorithms, we divide the whole dataset to have three independent partitions such  that training set has 90 subjects, the development set has 30 subjects and also the testing set has 30 subjects. Further, we perform the cross validation by randomising the selection of training, development and testing set for $N=5$ times and average the results, which are reported with standard deviation.

\section{Baseline Systems }
\label{sec:baseline}
In this section, we discuss baseline biometric systems used to benchmark the performance of biometric verification and presentation attack detection.
\subsection{Biometric verification }
\label{sec:BiometricBaselines}

\paragraph{Face Biometrics}
Two deep learning based face biometric systems are evaluated: VGG-Face\footnote{\mbox{Website: \url{www.robots.ox.ac.uk/~vgg/software/vgg_face}}} \cite{parkhi_deep_2015} and FaceNet \cite{schroff_facenet:_2015,sandberg_facenet_2017}.
The choice of networks is based on the obtained accuracy on the LFW dataset - challenging FR dataset \cite{LFWTech} where FaceNet reported an accuracy of 99.2\% and VGG-Face reported an accuracy of 98.95\%.

\paragraph{Eye Biometrics}
To evaluate the periocular biometric system, we have used five different methods that include: Coupled-Autoencoder \cite{ICIPCompitRaghu}, Collaborative representation of Deep-sparse Features \cite{ICIPCompitKiran}, Deep Convolution Neural Network (DCNN) features from pre-trained networks such as: AlexNet, VGG16 and  ResNet50. These baseline systems are selected based on the reported verification performance especially on the smartphone environment.

\paragraph{Audio-visual biometrics}
Inter-Session Variability (ISV) \cite{VOGT200817,McCool_IET_BMT_2013} and a deep learning based network (ResNet) \cite{le_robust_2018} is used for evaluating the voice biometrics.
ISV is a GMM-UBM \cite{reynolds_speaker_2000} based system which explicitly models the session variability.
We have used an extended ResNet implementation of \cite{le_robust_2018} named dilated residual network (DRN) which is publicly available\footnote{\url{https://www.idiap.ch/software/bob/docs/bob/bob.learn.pytorch/v0.0.4/guide_audio_extractor.html}}.
The DRN model is one of the state-of-the-art systems on the Voxceleb 1 database \cite{nagrani_voxceleb:_2017} evaluations achieving $4.8\%$ EER on the dataset.

The audio-visual system is the result of score fusion between the FaceNet and the DRN models which are the best performing algorithm for face and voice subsystems, respectively.

\subsection{ Presentation Attack Detection Algorithms}
\label{sec:PADBaselines}
\paragraph{Face and Eye Attack Instruments}
We report the detection performance of five different PAD algorithms on both eye and face attack instruments. The baseline algorithms include Local Binary Pattern (LBP) \cite{BioEvaSpoof}, Local Phase Quantisation (LPQ)\cite{BioEvaSpoof}, Binarised Statistical Image Features (BSIF)\cite{Raghusipco}, Image Distortion Analysis (IDA) \cite{IDMJain} and Color texture features \cite{ColorLBPfacePAD}. We use linear SVM as the classifier that is trained and tested with these features following different evaluation protocols as discussed in Section~\ref{sec:Proto}. We have selected these five PAD algorithms as the baseline methods to cover the spectrum of both micro-texture and image quality based attack detection techniques.

\paragraph{Audio-visual attack instruments}
We use three features: MFCC,SCFC,LFCC with two classifiers (SVM and GMM) \cite{sahidullah_comparison_2015} to develop voice PAD systems.
The same PAD systems that were used for still faces were also used here.
The final audio-visual PAD system is the result of score-level fusion of the best algorithms of face and audio PAD systems (Color Textures-SVM and LFCC-GMM).

\section{Experimental Results}
\label{sec:Exp}
In this section, we present the baseline algorithms performance evaluation on both biometric verification and presentation attack detection. The performance of the baseline algorithms is presented using the Equal Error Rate (\%) following the experimental protocol presented in Section \ref{sec:Proto}. The performance of the baseline PAD algorithms is presented using \textit{Bona fide Presentation Classification Error Rate (BPCER) and Attack Presentation Classification Error Rate (APCER)}. \textbf{BPCER} is defined as the proportion of bona fide presentations incorrectly classified as attacks while \textbf{APCER} is defined as the proportion of attack presentations incorrectly classified as bona fide presentations. In particular, we report the performance of the baseline PAD techniques by reporting the value of BPCER while fixing the APCER to 5\% and 10\% according to the recommendation from IS0/IEC 30107-3 \cite{ISO/IEC2015a}. The PAD evaluation protocol is presented in Section \ref{sec:Proto}.
\subsection{Biometric Verification Results}
\label{sec:ExpBIO}
Table \ref{tab:UniModalVerfi} indicates the performance of the uni-modal biometric systems with Eye, 2D Face and Audio-visual.  For the simplicity, we have used the self capture images from each session to present the quantitative results on eye verification using baseline methods (see Section \ref{sec:Proto} for evaluation protocols). Based on the obtained results as presented in Table \ref{tab:UniModalVerfi}, the following can be observed (1) among the five different methods, the DeepSparse-CRC \cite{ICIPCompitKiran} method shows the marginal improvement over other techniques in both Protocol-1  and Protocol-2 (c). (2) Both DeepSparse-CRC \cite{ICIPCompitKiran} and Deep Autoencoder \cite{ICIPCompitRaghu} algorithms indicate a degraded performance for Protocol-2 when compared to that of the Protocol-1. This can be attributed to the capture data quality reflected from all four sessions (S3-S6).

Both face biometric systems showed similar performance across all sessions.
The FaceNet system outperformed the VGG-Face system by a large margin.
However, the same FaceNet system's performance was degraded when evaluated on audio-visual data.
This can be attributed to the fact the camera was held at a reading position (lower down) during the audio-visual data capture process compared to holding device higher up when taking still face videos.
The DRN speaker verification system performed well compared to the ISV system (not shown for brevity, the ISV's EER on all sessions was $13.1\%$).
The worst performance was achieved on session 3 which can be mainly attributed to the noise (especially from wind) outdoor.
The final audio-visual biometric system was the average score of FaceNet and DRN systems.
Performing this fusion improved the results (except for session 4) which showed that the information from face and voice were mainly complementary in performing verification.

\begin{table}[htbp]
  \centering
  \caption{Baseline performance of uni-modal biometric system}
  \resizebox{1\linewidth}{!}{
    \begin{tabular}{|p{6.665em}|r|r|r|r|}
    \hline
    \textbf{Modality} & \multicolumn{1}{p{5em}|}{\textbf{Algorithms}} & \multicolumn{1}{p{5em}|}{\textbf{Enrolment}} & \multicolumn{1}{p{5em}|}{\textbf{Probe}} & \multicolumn{1}{p{5em}|}{\textbf{ EER(\%)}} \bigstrut\\
    \hline
    \multirow{25}[50]{*}{{Eye biometrics}} & \multicolumn{1}{c|}{\multirow{5}[10]{*}{Deep Autoencoder \cite{ICIPCompitRaghu} }} & \multicolumn{1}{c|}{\multirow{5}[10]{*}{Session 2}} & \multicolumn{1}{p{5em}|}{Session 3} & \multicolumn{1}{p{5em}|}{25.59} \bigstrut\\
\cline{4-5}    \multicolumn{1}{|c|}{} &       &       & \multicolumn{1}{p{5em}|}{Session 4} & \multicolumn{1}{p{5em}|}{23.88} \bigstrut\\
\cline{4-5}    \multicolumn{1}{|c|}{} &       &       & \multicolumn{1}{p{5em}|}{Session 5} & \multicolumn{1}{p{5em}|}{27.26} \bigstrut\\
\cline{4-5}    \multicolumn{1}{|c|}{} &       &       & \multicolumn{1}{p{5em}|}{Session 6} & \multicolumn{1}{p{5em}|}{23.07} \bigstrut\\
\cline{4-5}    \multicolumn{1}{|c|}{} &       &       & \multicolumn{1}{p{5em}|}{All Session} & \multicolumn{1}{p{5em}|}{25.67} \bigstrut\\
 \cline{2-5}    \multicolumn{1}{|c|}{} & \multicolumn{1}{c|}{\multirow{5}[10]{*}{DeepSparse-CRC \cite{ICIPCompitKiran} }} & \multicolumn{1}{c|}{\multirow{5}[10]{*}{Session 2}} & \multicolumn{1}{p{5em}|}{Session 3} & \multicolumn{1}{p{5em}|}{21.32} \bigstrut\\
\cline{4-5}    \multicolumn{1}{|c|}{} &       &       & \multicolumn{1}{p{5em}|}{Session 4} & \multicolumn{1}{p{5em}|}{22.30} \bigstrut\\
\cline{4-5}    \multicolumn{1}{|c|}{} &       &       & \multicolumn{1}{p{5em}|}{Session 5} & \multicolumn{1}{p{5em}|}{22.55} \bigstrut\\
\cline{4-5}    \multicolumn{1}{|c|}{} &       &       & \multicolumn{1}{p{5em}|}{Session 6} & \multicolumn{1}{p{5em}|}{23.54} \bigstrut\\
\cline{4-5}    \multicolumn{1}{|c|}{} &       &       & \multicolumn{1}{p{5em}|}{All Session} & \multicolumn{1}{p{5em}|}{22.41} \bigstrut\\
\cline{2-5}    \multicolumn{1}{|c|}{} & \multicolumn{1}{c|}{\multirow{5}[10]{*}{AlexNet features}} & \multicolumn{1}{c|}{\multirow{5}[10]{*}{Session 2}} & \multicolumn{1}{p{5em}|}{Session 3} & \multicolumn{1}{p{5em}|}{24.23} \bigstrut\\
\cline{4-5}    \multicolumn{1}{|c|}{} &       &       & \multicolumn{1}{p{5em}|}{Session 4} & \multicolumn{1}{p{5em}|}{28.72} \bigstrut\\
\cline{4-5}    \multicolumn{1}{|c|}{} &       &       & \multicolumn{1}{p{5em}|}{Session 5} & \multicolumn{1}{p{5em}|}{27.21} \bigstrut\\
\cline{4-5}    \multicolumn{1}{|c|}{} &       &       & \multicolumn{1}{p{5em}|}{Session 6} & \multicolumn{1}{p{5em}|}{21.46} \bigstrut\\
\cline{4-5}    \multicolumn{1}{|c|}{} &       &       & \multicolumn{1}{p{5em}|}{All Session} & \multicolumn{1}{p{5em}|}{25.46} \bigstrut\\
\cline{2-5}    \multicolumn{1}{|c|}{} & \multicolumn{1}{c|}{\multirow{5}[10]{*}{ResNet features}} & \multicolumn{1}{c|}{\multirow{5}[10]{*}{Session 2}} & \multicolumn{1}{p{5em}|}{Session 3} & \multicolumn{1}{p{5em}|}{29.59} \bigstrut\\
\cline{4-5}    \multicolumn{1}{|c|}{} &       &       & \multicolumn{1}{p{5em}|}{Session 4} & \multicolumn{1}{p{5em}|}{24.78} \bigstrut\\
\cline{4-5}    \multicolumn{1}{|c|}{} &       &       & \multicolumn{1}{p{5em}|}{Session 5} & \multicolumn{1}{p{5em}|}{24.17} \bigstrut\\
\cline{4-5}    \multicolumn{1}{|c|}{} &       &       & \multicolumn{1}{p{5em}|}{Session 6} & \multicolumn{1}{p{5em}|}{20.79} \bigstrut\\
\cline{4-5}    \multicolumn{1}{|c|}{} &       &       & \multicolumn{1}{p{5em}|}{All Session} & \multicolumn{1}{p{5em}|}{24.77} \bigstrut\\
\cline{2-5}    \multicolumn{1}{|c|}{} & \multicolumn{1}{c|}{\multirow{5}[10]{*}{VGG16 features}} & \multicolumn{1}{c|}{\multirow{5}[10]{*}{Session 2}} & \multicolumn{1}{p{5em}|}{Session 3} & \multicolumn{1}{p{5em}|}{34.62} \bigstrut\\
\cline{4-5}    \multicolumn{1}{|c|}{} &       &       & \multicolumn{1}{p{5em}|}{Session 4} & \multicolumn{1}{p{5em}|}{27.41} \bigstrut\\
\cline{4-5}    \multicolumn{1}{|c|}{} &       &       & \multicolumn{1}{p{5em}|}{Session 5} & \multicolumn{1}{p{5em}|}{27.45} \bigstrut\\
\cline{4-5}    \multicolumn{1}{|c|}{} &       &       & \multicolumn{1}{p{5em}|}{Session 6} & \multicolumn{1}{p{5em}|}{24.82} \bigstrut\\
\cline{4-5}    \multicolumn{1}{|c|}{} &       &       & \multicolumn{1}{p{5em}|}{All Session} & \multicolumn{1}{p{5em}|}{28.62} \bigstrut\\
    \hline
    \multirow{10}[50]{*}{{Face biometrics}} & \multicolumn{1}{c|}{\multirow{5}[10]{*}{FaceNet (InceptionResNetV1) \cite{sandberg_facenet_2017} }} & \multicolumn{1}{c|}{\multirow{5}[10]{*}{Session 2}} & \multicolumn{1}{p{5em}|}{Session 3} & \multicolumn{1}{p{5em}|}{4.3} \bigstrut\\
\cline{4-5}    \multicolumn{1}{|c|}{} &       &       & \multicolumn{1}{p{5em}|}{Session 4} & \multicolumn{1}{p{5em}|}{3.3} \bigstrut\\
\cline{4-5}    \multicolumn{1}{|c|}{} &       &       & \multicolumn{1}{p{5em}|}{Session 5} & \multicolumn{1}{p{5em}|}{5.3} \bigstrut\\
\cline{4-5}    \multicolumn{1}{|c|}{} &       &       & \multicolumn{1}{p{5em}|}{Session 6} & \multicolumn{1}{p{5em}|}{2.7} \bigstrut\\
\cline{4-5}    \multicolumn{1}{|c|}{} &       &       & \multicolumn{1}{p{5em}|}{All Session} & \multicolumn{1}{p{5em}|}{4.2} \bigstrut\\
 \cline{2-5}    \multicolumn{1}{|c|}{} & \multicolumn{1}{c|}{\multirow{5}[10]{*}{VGG-Face \cite{VGG}}} & \multicolumn{1}{c|}{\multirow{5}[10]{*}{Session 2}} & \multicolumn{1}{p{5em}|}{Session 3} & \multicolumn{1}{p{5em}|}{17.2} \bigstrut\\
\cline{4-5}    \multicolumn{1}{|c|}{} &       &       & \multicolumn{1}{p{5em}|}{Session 4} & \multicolumn{1}{p{5em}|}{17.3} \bigstrut\\
\cline{4-5}    \multicolumn{1}{|c|}{} &       &       & \multicolumn{1}{p{5em}|}{Session 5} & \multicolumn{1}{p{5em}|}{16.6} \bigstrut\\
\cline{4-5}    \multicolumn{1}{|c|}{} &       &       & \multicolumn{1}{p{5em}|}{Session 6} & \multicolumn{1}{p{5em}|}{17.2} \bigstrut\\
\cline{4-5}    \multicolumn{1}{|c|}{} &       &       & \multicolumn{1}{p{5em}|}{All Session} & \multicolumn{1}{p{5em}|}{17.0} \bigstrut\\
    \hline
    \multirow{15}[50]{*}{{Audio-Visual}} & \multicolumn{1}{c|}{\multirow{5}[10]{*}{FaceNet (InceptionResNetV1) \cite{sandberg_facenet_2017}}} & \multicolumn{1}{c|}{\multirow{5}[10]{*}{Session 2}} & \multicolumn{1}{p{5em}|}{Session 3} & \multicolumn{1}{p{5em}|}{14.2} \bigstrut\\
\cline{4-5}    \multicolumn{1}{|c|}{} &       &       & \multicolumn{1}{p{5em}|}{Session 4} & \multicolumn{1}{p{5em}|}{13.1} \bigstrut\\
\cline{4-5}    \multicolumn{1}{|c|}{} &       &       & \multicolumn{1}{p{5em}|}{Session 5} & \multicolumn{1}{p{5em}|}{13.9} \bigstrut\\
\cline{4-5}    \multicolumn{1}{|c|}{} &       &       & \multicolumn{1}{p{5em}|}{Session 6} & \multicolumn{1}{p{5em}|}{13.0} \bigstrut\\
\cline{4-5}    \multicolumn{1}{|c|}{} &       &       & \multicolumn{1}{p{5em}|}{All Session} & \multicolumn{1}{p{5em}|}{13.5} \bigstrut\\
 \cline{2-5}    \multicolumn{1}{|c|}{} & \multicolumn{1}{c|}{\multirow{5}[10]{*}{DRN }} & \multicolumn{1}{c|}{\multirow{5}[10]{*}{Session 2}} & \multicolumn{1}{p{5em}|}{Session 3} & \multicolumn{1}{p{5em}|}{4.3} \bigstrut\\
\cline{4-5}    \multicolumn{1}{|c|}{} &       &       & \multicolumn{1}{p{5em}|}{Session 4} & \multicolumn{1}{p{5em}|}{2.1} \bigstrut\\
\cline{4-5}    \multicolumn{1}{|c|}{} &       &       & \multicolumn{1}{p{5em}|}{Session 5} & \multicolumn{1}{p{5em}|}{3.2} \bigstrut\\
\cline{4-5}    \multicolumn{1}{|c|}{} &       &       & \multicolumn{1}{p{5em}|}{Session 6} & \multicolumn{1}{p{5em}|}{3.3} \bigstrut\\
\cline{4-5}    \multicolumn{1}{|c|}{} &       &       & \multicolumn{1}{p{5em}|}{All Session} & \multicolumn{1}{p{5em}|}{3.2} \bigstrut\\
 \cline{2-5}    \multicolumn{1}{|c|}{} & \multicolumn{1}{c|}{\multirow{5}[10]{*}{FaceNet-DRN-score-mean-fusion}} & \multicolumn{1}{c|}{\multirow{5}[10]{*}{Session 2}} & \multicolumn{1}{p{5em}|}{Session 3} & \multicolumn{1}{p{5em}|}{3.4} \bigstrut\\
\cline{4-5}    \multicolumn{1}{|c|}{} &       &       & \multicolumn{1}{p{5em}|}{Session 4} & \multicolumn{1}{p{5em}|}{3.0} \bigstrut\\
\cline{4-5}    \multicolumn{1}{|c|}{} &       &       & \multicolumn{1}{p{5em}|}{Session 5} & \multicolumn{1}{p{5em}|}{3.1} \bigstrut\\
\cline{4-5}    \multicolumn{1}{|c|}{} &       &       & \multicolumn{1}{p{5em}|}{Session 6} & \multicolumn{1}{p{5em}|}{2.9} \bigstrut\\
\cline{4-5}    \multicolumn{1}{|c|}{} &       &       & \multicolumn{1}{p{5em}|}{All Session} & \multicolumn{1}{p{5em}|}{3.1} \bigstrut\\
    \hline
    \end{tabular}%
    }
  \label{tab:UniModalVerfi}%
\end{table}%

\subsection{Biometric vulnerability assessment}
\label{sec:SWANVul}
\begin{figure}[htb]
    \centering
    \includegraphics[width=\columnwidth]{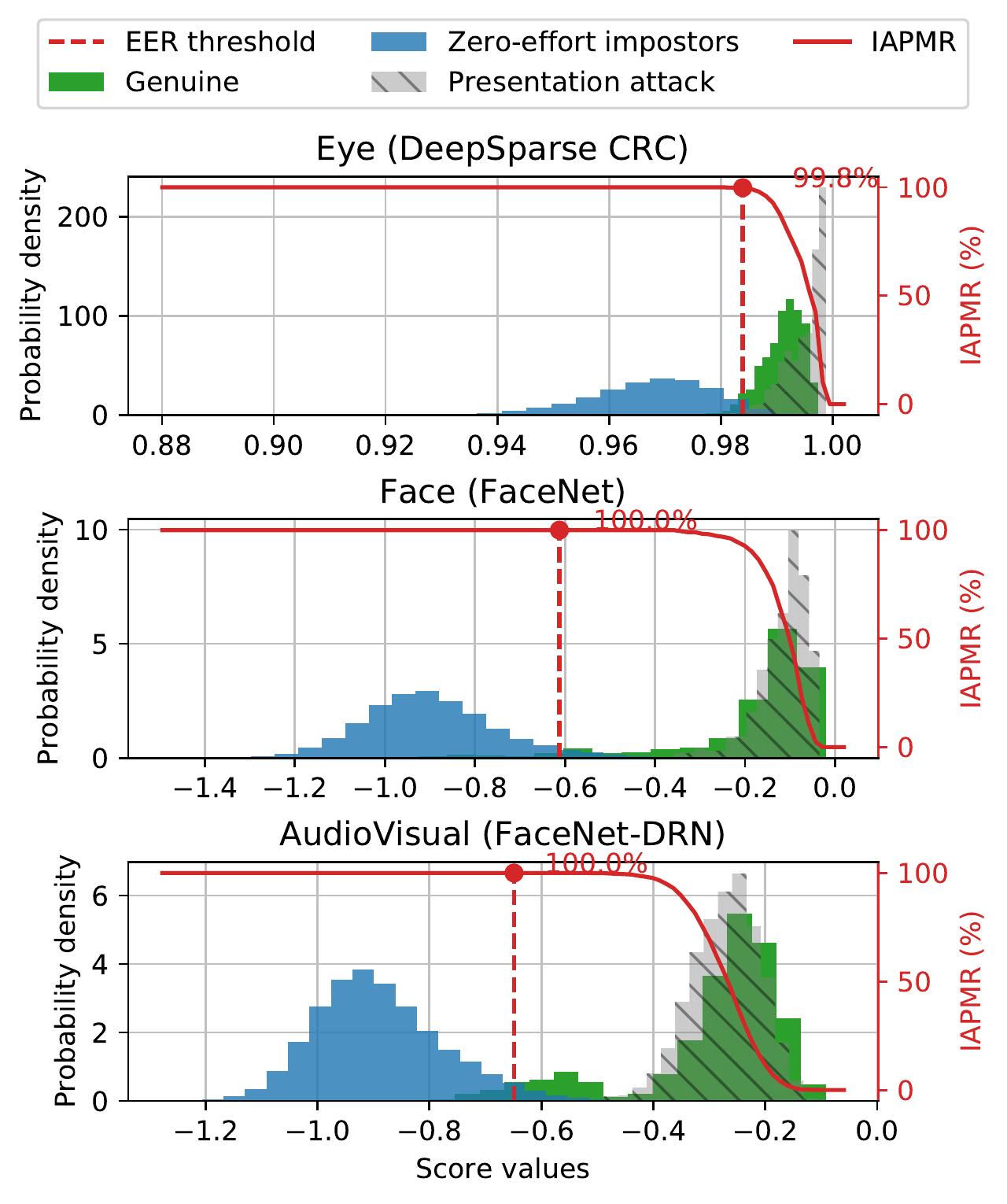}
    \caption{Vulnerability analysis of uni-modal biometric systems }
    \label{fig:vuln-hist-modalities}
\end{figure}

Figure~\ref{fig:vuln-hist-modalities} illustrates the vulnerability analysis on the uni-modal biometrics from the SWAN multimodal biometric dataset. The vulnerability analysis is performed using the baseline uni-modal biometric in which, the bona fide samples are enrolled and presentation attack samples are used as the probe. Finally, the comparisons scores obtained on the probe samples are compared against the operational threshold set, and finally, the quantitative results of the vulnerability are presented using an Impostor Attack Presentation Match Rate (IAPMR (\%)).  For vulnerability analysis with eye biometric system we have used the DeepSparse-CRC \cite{ICIPCompitKiran} system, for 2D face biometrics we have used FaceNet (InceptionResNetV1) \cite{sandberg_facenet_2017} and for audio-visual biometric system we have employed FaceNet (InceptionResNetV1) \cite{sandberg_facenet_2017} for face and DRN \cite{le_robust_2018} for voice whose scores are fused at comparisons core level.  All three types of presentation attack samples are used as the probe samples to compute the IAMPR. As indicated in the Figure~\ref{fig:vuln-hist-modalities}, all three modalities have indicated IAMPR of $100\%$ on both face and audiovisual biometrics and $99.8\%$ on eye biometrics. The obtained results further justify the quality of the presentation attacks generated in this work and also the need for Presentation Attack Detection (PAD)  techniques to mitigate the presentation attacks.

\subsection{Biometric PAD results}
\label{sec:PADResults}

\begin{table}[htbp]
  \centering
  \caption{Baseline performance of PAD techniques on Eye biometrics}
    \resizebox{1\linewidth}{!}{
    \begin{tabular}{|c|p{7.25em}|p{5.915em}|p{5.915em}|p{6.585em}|p{6.75em}|}
    \hline
    \multicolumn{1}{|c|}{\multirow{3}[6]{*}{\textbf{Evaluation Protocol}}} & \multirow{3}[6]{*}{\textbf{Algorithms}} & \textbf{Development Database} & \multicolumn{3}{p{17.25em}|}{\textbf{Testing Database}} \bigstrut\\
\cline{3-6}          & \multicolumn{1}{c|}{} & \multirow{2}[4]{*}{\textbf{D-EER(\%)}} & \multirow{2}[4]{*}{\textbf{D-EER(\%)}} & \multicolumn{2}{p{11.335em}|}{\textbf{BPCER@ APCER}} \bigstrut\\
\cline{5-6}          & \multicolumn{1}{c|}{} & \multicolumn{1}{c|}{} & \multicolumn{1}{c|}{} & \textbf{= 5(\%)} & \textbf{= 10(\%)} \bigstrut\\
    \hline
    \multicolumn{1}{|c|}{\multirow{5}[10]{*}{Protocol-1 (PA.E.1)}} & \textbf{BSIF-SVM \cite{Raghusipco}} & 10.14 $\pm$ 1.16 & 11.78 $\pm$2.62 & 57.95 $\pm$ 40.16 & 37.54 $\pm$ 29.97 \bigstrut\\
\cline{2-6}          & \textbf{Color Textures-SVM \cite{ColorLBPfacePAD}} & 3.65 $\pm$ 2.13 & 1.07 $\pm$ 1.47 & 17.59 $\pm$ 16.49 & 10.51$\pm$11.54 \bigstrut\\
\cline{2-6}          & \textbf{IDA-SVM \cite{IDMJain}} & 19.42 $\pm$ 17.46 & 24.23 $\pm$ 19.80 & 44.87 $\pm$ 26.86 & 37.77 $\pm$ 25.83 \bigstrut\\
\cline{2-6}          & \textbf{LBP-SVM \cite{BioEvaSpoof}} & 5.52 $\pm$ 1.17 & 9.55 $\pm$ 2.15 & 17.70 $\pm$ 14.90 & 16.86 $\pm$ 16.41 \bigstrut\\
\cline{2-6}          & \textbf{LPQ-SVM \cite{BioEvaSpoof}} & 13.14 $\pm$ 3.74 & 12.63 $\pm$ 2.99 & 53.80 $\pm$ 32.60 & 37.34 $\pm$ 28.74 \bigstrut\\
    \hline
    \multicolumn{1}{|c|}{\multirow{5}[10]{*}{Protocol-1 (PA.E.4)}} & \textbf{BSIF-SVM \cite{Raghusipco}} & 7.51 $\pm$ 2.04 & 8.53 $\pm$ 1.92 & 34.41 $\pm$ 13.80 & 18.82 $\pm$ 5.59 \bigstrut\\
\cline{2-6}          & \textbf{Color Textures-SVM \cite{ColorLBPfacePAD}} & 0.06 $\pm$ 0.06 & 0.85 $\pm$ 0.57 & 0.31 $\pm$ 0.58 & 0.11+ 0.22 \bigstrut\\
\cline{2-6}          & \textbf{IDA-SVM \cite{IDMJain}} & 20.41 $\pm$ 8.24 & 24.35 $\pm$ 13.09 & 61.88 $\pm$ 32.19 & 55.44 $\pm$ 37.87 \bigstrut\\
\cline{2-6}          & \textbf{LBP-SVM \cite{BioEvaSpoof}} & 4.47 $\pm$ 0.92 & 7.26 $\pm$ 3.31 & 8.13 $\pm$ 7.44 & 5.51 $\pm$ 6.74 \bigstrut\\
\cline{2-6}          & \textbf{LPQ-SVM \cite{BioEvaSpoof}} & 5.22 $\pm$ 1.44 & 5.18 $\pm$ 1.25 & 22.07 $\pm$ 10.27 & 9.84 $\pm$ 6.54 \bigstrut\\
    \hline
    \multicolumn{1}{|c|}{\multirow{5}[10]{*}{Protocol-1 (PA.E.5)}} & \textbf{BSIF-SVM \cite{Raghusipco}} & 10.33 $\pm$ 2.03 & 13.60 $\pm$ 2.98 & 31.96 $\pm$ 17.10 & 21.80 $\pm$ 11.40 \bigstrut\\
\cline{2-6}          & \textbf{Color Textures-SVM \cite{ColorLBPfacePAD}} & 1.35 $\pm$ 1.33 & 0.97 $\pm$ 0.52 & 1.64 $\pm$ 3.49 & 0.76 $\pm$ 1.66 \bigstrut\\
\cline{2-6}          & \textbf{IDA-SVM \cite{IDMJain}} & 17.81 $\pm$ 9.22 & 21.50 $\pm$ 6.89 & 29.18 $\pm$ 19.26 & 20.97 $\pm$ 14.33 \bigstrut\\
\cline{2-6}          & \textbf{LBP-SVM \cite{BioEvaSpoof}} & 1.12 $\pm$ 0.48 & 1.52 $\pm$ 0.65 & 2.14 $\pm$ 2.13 & 1.02 $\pm$ 1.24 \bigstrut\\
\cline{2-6}          & \textbf{LPQ-SVM \cite{BioEvaSpoof}} & 1.34 $\pm$ 0.65 & 1.91 $\pm$ 0.29 & 11.65 $\pm$ 24.25 & 4.76 $\pm$ 10.35 \bigstrut\\
    \hline
    \multicolumn{1}{|c|}{\multirow{5}[10]{*}{\textbf{Protocol-2}}} & \textbf{BSIF-SVM \cite{Raghusipco}} & 10.02 $\pm$ 1.34 & 12.04 $\pm$ 2.89 & 42.02 $\pm$ 21.27 & 24.60 $\pm$ 10.51 \bigstrut\\
\cline{2-6}          & \textbf{Color Textures-SVM \cite{ColorLBPfacePAD}} & 2.01 $\pm$ 0.98 & 2.68 $\pm$ 0.72 & 4.79 $\pm$ 8.37 & 1.93 $\pm$ 3.50 \bigstrut\\
\cline{2-6}          & \textbf{IDA-SVM \cite{IDMJain}} & 35.40 $\pm$ 9.78 & 38.16 $\pm$ 8.37 & 80.35 $\pm$ 13.96 & 73.29 $\pm$ 16.75 \bigstrut\\
\cline{2-6}          & \textbf{LBP-SVM \cite{BioEvaSpoof}} & 6.34 $\pm$ 0.63 & 8.95 $\pm$ 2.11 & 22.02 $\pm$ 25.93 & 14.75 $\pm$ 19.41 \bigstrut\\
\cline{2-6}          & \textbf{LPQ-SVM \cite{BioEvaSpoof}} & 6.92 $\pm$ 2.77 & 7.20 $\pm$  1.73 & 22.69 $\pm$ 12.80 & 9.57 $\pm$ 5.79 \bigstrut\\
    \hline
    \end{tabular}%
    }
  \label{tab:EyePAD}%
\end{table}%

\begin{table}[htbp]
  \centering
  \caption{Baseline performance of PAD techniques on face biometrics}
    \resizebox{1\linewidth}{!}{
    \begin{tabular}{|c|p{6.915em}|p{5.665em}|p{5.5em}|p{6.335em}|p{6.165em}|}
    \hline
    \multicolumn{1}{|c|}{\multirow{3}[6]{*}{\textbf{Evaluation Protocol}}} & \multirow{3}[6]{*}{\textbf{Algorithms}} & \textbf{Development Database} & \multicolumn{3}{p{18em}|}{\textbf{Testing Database}} \bigstrut\\
\cline{3-6}          & \multicolumn{1}{c|}{} & \multirow{2}[4]{*}{\textbf{D-EER(\%)}} & \multirow{2}[4]{*}{\textbf{D-EER(\%)}} & \multicolumn{2}{p{12.5em}|}{\textbf{BPCER@ APCER}} \bigstrut\\
\cline{5-6}          & \multicolumn{1}{c|}{} & \multicolumn{1}{c|}{} & \multicolumn{1}{c|}{} & \textbf{=5(\%)} & \textbf{=10(\%)} \bigstrut\\
    \hline
    \multicolumn{1}{|c|}{\multirow{5}[10]{*}{\textbf{Protocol-1 (PA.F.1)}}} & \textbf{BSIF-SVM \cite{Raghusipco}} & 5.20$\pm$8.75 & 10.13$\pm$2.58 & 38.92$\pm$29.24 & 30.16$\pm$ 24.20 \bigstrut\\
\cline{2-6}          & \textbf{Color Textures-SVM \cite{ColorLBPfacePAD}} & 2.20$\pm$1.25 & 8.71$\pm$1.28 & 24.87$\pm$ 29.68 & 18.90$\pm$22.16 \bigstrut\\
\cline{2-6}          & \textbf{IDA-SVM \cite{IDMJain}} & 34.50$\pm$22.41 & 37.61$\pm$22.32 & 63.78$\pm$30.58 & 56.77$\pm$36.27 \bigstrut\\
\cline{2-6}          & \textbf{LBP-SVM \cite{BioEvaSpoof}} & 3.08$\pm$1.91 & 13.63$\pm$2.06 & 39.23$\pm$27.97 & 33.78$\pm$24.86 \bigstrut\\
\cline{2-6}          & \textbf{LPQ-SVM \cite{BioEvaSpoof}} & 5.67$\pm$2.67 & 13.87$\pm$2.69 & 30.52$\pm$12.52 & 22.54$\pm$9.47 \bigstrut\\
    \hline
    \multicolumn{1}{|c|}{\multirow{5}[10]{*}{\textbf{Protocol-1 (PA.F.5)}}} & \textbf{BSIF-SVM \cite{Raghusipco}} & 17.16$\pm$5.11 & 20.30$\pm$1.19 & 67.48$\pm$30.14 & 50.32$\pm$23.73 \bigstrut\\
\cline{2-6}          & \textbf{Color Textures-SVM \cite{ColorLBPfacePAD}} & 11.57$\pm$4.37 & 12.46$\pm$3.21 & 45.76$\pm$19.67 & 27.96$\pm$9.17 \bigstrut\\
\cline{2-6}          & \textbf{IDA-SVM \cite{IDMJain}} & 24.55$\pm$3.11 & 29.08$\pm$13.40 & 56.53$\pm$26.66 & 46.87$\pm$28.95 \bigstrut\\
\cline{2-6}          & \textbf{LBP-SVM \cite{BioEvaSpoof}} & 15.83$\pm$3.65 & 17.13$\pm$4.38 & 42.05$\pm$18.31 & 30.96$\pm$16.13 \bigstrut\\
\cline{2-6}          & \textbf{LPQ-SVM \cite{BioEvaSpoof}} & 24.25$\pm$2.75 & 18.15$\pm$3.07 & 63.63$\pm$25.64 & 53.83$\pm$26.49 \bigstrut\\
    \hline
    \multicolumn{1}{|c|}{\multirow{5}[10]{*}{\textbf{Protocol-1 (PA.F.6)}}} & \textbf{BSIF-SVM \cite{Raghusipco}} & 17.76$\pm$4.27 & 27.74$\pm$1.55 & 70.50$\pm$16.30 & 59.48$\pm$15.43 \bigstrut\\
\cline{2-6}          & \textbf{Color Textures-SVM \cite{ColorLBPfacePAD}} & 4.21$\pm$1.79 & 5.66$\pm$2.13 & 12.29$\pm$7.79 & 8.08$\pm$4.85 \bigstrut\\
\cline{2-6}          & \textbf{IDA-SVM \cite{IDMJain}} & 31.78$\pm$12.69 & 30.99$\pm$12.99 & 71.09$\pm$29.95 & 62.75$\pm$37.14 \bigstrut\\
\cline{2-6}          & \textbf{LBP-SVM \cite{BioEvaSpoof}} & 6.70$\pm$0.98 & 9.41$\pm$2.59 & 32.21$\pm$22.19 & 19.30$\pm$16.50 \bigstrut\\
\cline{2-6}          & \textbf{LPQ-SVM \cite{BioEvaSpoof}} & 17.07$\pm$3.18 & 17.04$\pm$2.41 & 51.63$\pm$26.95 & 38.36$\pm$25.62 \bigstrut\\
    \hline
    \multicolumn{1}{|c|}{\multirow{5}[10]{*}{\textbf{Protocol-2}}} & \textbf{BSIF-SVM \cite{Raghusipco}} & 20.05$\pm$3.97 & 25.55$\pm$1.51 & 58.12$\pm$13.61 & 46.06$\pm$12.76 \bigstrut\\
\cline{2-6}          & \textbf{Color Textures-SVM \cite{ColorLBPfacePAD}} & 7.07$\pm$1.13 & 14.45$\pm$1.56 & 47.31$\pm$7.66 & 36.06$\pm$6.33 \bigstrut\\
\cline{2-6}          & \textbf{IDA-SVM \cite{IDMJain}} & 32.66$\pm$19.22 & 33.49$\pm$16.17 & 66.54$\pm$15.45 & 58.79$\pm$16.99 \bigstrut\\
\cline{2-6}          & \textbf{LBP-SVM \cite{BioEvaSpoof}} & 27.67$\pm$11.90 & 35.95$\pm$7.65 & 75.09$\pm$12.18 & 64.23$\pm$16.14 \bigstrut\\
\cline{2-6}          & \textbf{LPQ-SVM \cite{BioEvaSpoof}} & 18.47$\pm$3.93 & 20.23$\pm$3.84 & 48.28$\pm$10.95 & 39.05$\pm$9.33 \bigstrut\\
    \hline
    \end{tabular}%
    }
  \label{tab:FacePAD}%
\end{table}%

\begin{table}[htbp]
\caption{Baseline performance of PAD techniques on Talking Faces biometrics}
\label{tab:TalkingFacePAD}
\resizebox{\columnwidth}{!}{%
\begin{tabular}{|l|l|l|l|l|l|}
\hline
\multirow{3}{*}{\textbf{Evaluation Protocol}} & \multirow{3}{*}{\textbf{Algorithms}} & \textbf{\begin{tabular}[c]{@{}l@{}}Development\\ Database\end{tabular}} & \multicolumn{3}{l|}{\textbf{Testing Database}}                                                         \\ \cline{3-6}
                                              &                                      & \multirow{2}{*}{\textbf{D-EER (\%)}}                                    & \multicolumn{1}{c|}{\multirow{2}{*}{\textbf{D-EER (\%)}}} & \multicolumn{2}{l|}{\textbf{BPCER@ APCER}} \\ \cline{5-6}
                                              &                                      &                                                                         & \multicolumn{1}{c|}{}                                     & \textbf{=5(\%)}     & \textbf{=10(\%)}     \\ \hline
\multirow{5}{*}{\textbf{Protocol-1 (PA.F.5)}} & \textbf{BSIF-SVM}                    & 23.25$\pm$5.08                                                          & 27.32$\pm$3.91                                            & 97.92$\pm$1.77      & 91.88$\pm$6.18       \\ \cline{2-6}
                                              & \textbf{Color Textures-SVM}          & 2.67$\pm$1.57                                                           & 5.45$\pm$1.50                                             & 57.93$\pm$27.78     & 43.67$\pm$27.25      \\ \cline{2-6}
                                              & \textbf{IDA-SVM}                     & 35.00$\pm$8.99                                                          & 37.87$\pm$9.20                                            & 53.79$\pm$44.49     & 50.84$\pm$43.82      \\ \cline{2-6}
                                              & \textbf{LBP-SVM}                     & 17.83$\pm$12.12                                                         & 24.06$\pm$11.01                                           & 52.37$\pm$37.12     & 39.69$\pm$33.97      \\ \cline{2-6}
                                              & \textbf{LPQ-SVM}                     & 5.25$\pm$2.86                                                           & 8.66$\pm$1.38                                             & 30.87$\pm$25.64     & 19.46$\pm$19.70      \\ \hline
\multirow{5}{*}{\textbf{Protocol-1 (PA.F.6)}} & \textbf{BSIF-SVM}                    & 21.74$\pm$4.55                                                          & 29.34$\pm$3.67                                            & 81.30$\pm$11.92     & 68.47$\pm$19.95      \\ \cline{2-6}
                                              & \textbf{Color Textures-SVM}          & 5.47$\pm$2.98                                                           & 15.21$\pm$2.86                                            & 48.55$\pm$40.99     & 43.67$\pm$43.22      \\ \cline{2-6}
                                              & \textbf{IDA-SVM}                     & 37.88$\pm$17.72                                                         & 41.69$\pm$9.97                                            & 69.82$\pm$35.84     & 66.42$\pm$35.48      \\ \cline{2-6}
                                              & \textbf{LBP-SVM}                     & 38.51$\pm$8.25                                                          & 41.13$\pm$8.85                                            & 96.36$\pm$4.05      & 92.79$\pm$7.52       \\ \cline{2-6}
                                              & \textbf{LPQ-SVM}                     & 19.17$\pm$3.06                                                          & 25.10$\pm$1.97                                            & 63.90$\pm$33.95     & 54.94$\pm$37.27      \\ \hline
\multirow{5}{*}{\textbf{Protocol-2}}          & \textbf{BSIF-SVM}                    & 26.48$\pm$3.31                                                          & 24.56$\pm$2.95                                            & 65.17$\pm$14.69     & 53.78$\pm$17.40      \\ \cline{2-6}
                                              & \textbf{Color Textures-SVM}          & 4.31$\pm$1.50                                                           & 5.18$\pm$1.58                                             & 9.72$\pm$7.24       & 5.22$\pm$4.61        \\ \cline{2-6}
                                              & \textbf{IDA-SVM}                     & 37.07$\pm$10.02                                                         & 39.15$\pm$9.48                                            & 65.94$\pm$35.50     & 58.39$\pm$36.43      \\ \cline{2-6}
                                              & \textbf{LBP-SVM}                     & 34.00$\pm$7.64                                                          & 31.35$\pm$6.93                                            & 52.11$\pm$38.95     & 45.56$\pm$41.87      \\ \cline{2-6}
                                              & \textbf{LPQ-SVM}                     & 16.72$\pm$1.98                                                          & 16.86$\pm$3.68                                            & 60.44$\pm$21.89     & 51.61$\pm$25.25      \\ \hline
\end{tabular}%
}
\end{table}

\begin{table}[htbp]
\caption{Baseline performance of PAD techniques on Voice biometrics}
\label{tab:VoicePAD}
\resizebox{\columnwidth}{!}{%
\begin{tabular}{|l|l|l|l|l|l|}
\hline
\multirow{3}{*}{\textbf{Evaluation Protocol}} & \multirow{3}{*}{\textbf{Algorithms}} & \textbf{\begin{tabular}[c]{@{}l@{}}Development\\ Database\end{tabular}} & \multicolumn{3}{l|}{\textbf{Testing Database}}                                                         \\ \cline{3-6}
                                              &                                      & \multirow{2}{*}{\textbf{D-EER (\%)}}                                    & \multicolumn{1}{c|}{\multirow{2}{*}{\textbf{D-EER (\%)}}} & \multicolumn{2}{l|}{\textbf{BPCER@ APCER}} \\ \cline{5-6}
                                              &                                      &                                                                         & \multicolumn{1}{c|}{}                                     & \textbf{=5(\%)}     & \textbf{=10(\%)}     \\ \hline
\multirow{3}{*}{\textbf{Protocol-1 (PA.V.4)}} & \textbf{MFCC-SVM}                    & 0.00$\pm$0.00                                                           & 0.11$\pm$0.13                                             & 0.00$\pm$0.00       & 0.00$\pm$0.00        \\ \cline{2-6}
                                              & \textbf{LFCC-GMM}                    & 0.00$\pm$0.00                                                           & 0.00$\pm$0.00                                             & 0.00$\pm$0.00       & 0.00$\pm$0.00        \\ \cline{2-6}
                                              & \textbf{SCFC-GMM}                    & 1.28$\pm$0.38                                                           & 1.21$\pm$0.38                                             & 0.02$\pm$0.03       & 0.00$\pm$0.00        \\ \hline
\multirow{3}{*}{\textbf{Protocol-1 (PA.V.7)}} & \textbf{MFCC-SVM}                    & 0.00$\pm$0.00                                                           & 0.12$\pm$0.10                                             & 0.00$\pm$0.00       & 0.00$\pm$0.00        \\ \cline{2-6}
                                              & \textbf{LFCC-GMM}                    & 0.00$\pm$0.00                                                           & 0.00$\pm$0.00                                             & 0.00$\pm$0.00       & 0.00$\pm$0.00        \\ \cline{2-6}
                                              & \textbf{SCFC-GMM}                    & 1.58$\pm$0.49                                                           & 1.17$\pm$0.21                                             & 0.08$\pm$0.06       & 0.00$\pm$0.00        \\ \hline
\multirow{3}{*}{\textbf{Protocol-2}}          & \textbf{MFCC-SVM}                    & 1.52$\pm$0.43                                                           & 1.67$\pm$0.17                                             & 0.14$\pm$0.16       & 0.03$\pm$0.04        \\ \cline{2-6}
                                              & \textbf{LFCC-GMM}                    & 0.00$\pm$0.00                                                           & 0.00$\pm$0.00                                             & 0.00$\pm$0.00       & 0.00$\pm$0.00        \\ \cline{2-6}
                                              & \textbf{SCFC-GMM}                    & 2.09$\pm$0.41                                                           & 1.48$\pm$0.07                                             & 0.04$\pm$0.02       & 0.00$\pm$0.00        \\ \hline
\end{tabular}%
}
\end{table}

\begin{table}[htbp]
\caption{Baseline performance of PAD techniques on AudioVisual biometrics}
\label{tab:AudioVisualPAD}
\resizebox{\columnwidth}{!}{%
\begin{tabular}{|l|l|l|l|l|l|}
\hline
\multirow{3}{*}{\textbf{Evaluation Protocol}}                                                  & \multirow{3}{*}{\textbf{Algorithm:}}                                                                                    & \textbf{\begin{tabular}[c]{@{}l@{}}Development\\ Database\end{tabular}} & \multicolumn{3}{l|}{\textbf{Testing Database}}                                                         \\ \cline{3-6}
                                                                                               &                                                                                                                         & \multirow{2}{*}{\textbf{D-EER (\%)}}                                    & \multicolumn{1}{c|}{\multirow{2}{*}{\textbf{D-EER (\%)}}} & \multicolumn{2}{l|}{\textbf{BPCER@ APCER}} \\ \cline{5-6}
                                                                                               &                                                                                                                         &                                                                         & \multicolumn{1}{c|}{}                                     & \textbf{=5(\%)}      & \textbf{=10(\%)}    \\ \hline
\textbf{\begin{tabular}[c]{@{}l@{}}Protocol-1 Low-Quality\\ (PA.F.5 and PA.V.7)\end{tabular}}  & \multirow{3}{*}{\textbf{\begin{tabular}[c]{@{}l@{}}Fusion of \\ Color Textures-SVM \\ and\\ LFCC-GMM\end{tabular}}} & 0.08$\pm$0.17                                                           & 0.09$\pm$0.11                                             & 0.65$\pm$0.68        & 0.24$\pm$0.15       \\ \cline{1-1} \cline{3-6}
\textbf{\begin{tabular}[c]{@{}l@{}}Protocol-1 High-Quality\\ (PA.F.6 and PA.V.4)\end{tabular}} &                                                                                                                         & 21.74$\pm$4.55                                                          & 29.34$\pm$3.67  8                                         & 1.30$\pm$11.92  6    & 8.47$\pm$19.95      \\ \cline{1-1} \cline{3-6}
\textbf{\begin{tabular}[c]{@{}l@{}}Protocol-2\\ All Attacks\end{tabular}}                      &                                                                                                                         & 26.48$\pm$3.31                                                          & 24.56$\pm$2.95  6                                         & 5.17$\pm$14.69  5    & 3.78$\pm$17.40      \\ \hline
\end{tabular}%
}
\end{table}

Table \ref{tab:EyePAD} indicates the quantitative performance of the baseline PAD algorithms on the eye recognition subsystem on Protocol-1 and Protocol-2. Based on the obtained results, the following can be observed: (1) Among the three different PAI, the detection of PA.E.4 (iPad Pro front camera) indicates an excellent detection accuracy on all five baseline methods employed in this work. (2) Among five different baseline PAD algorithms, the PAD technique based on Color Textures-SVM \cite{ColorLBPfacePAD} has indicated the best performance on both Protocol-1 and Protocol-2. (3) The performance of the PAD algorithms in Protocol-2 shows a degraded performance when compared to that of the Protocol-1.

Table \ref{tab:FacePAD} and \ref{tab:TalkingFacePAD} indicates the quantitative performance of the baseline PAD algorithms on 2D face modality on Protocol-1 and Protocol-2. Based on the obtained results, it is noted that (1) the performance of the PAD techniques are degraded in the Protocol-2 when compared to Protocol-1. (2) Among five different baseline PAD algorithms, the PAD technique based on Color Textures-SVM \cite{ColorLBPfacePAD} has indicated the best performance on both Protocol-1 and Protocol-2.

Table \ref{tab:VoicePAD} indicates the quantitative performance of the baseline PAD algorithms on voice biometrics. In all protocols the LFCC-GMM system showed well performance ($0\%$ error rates) in detection of the presentation attacks.
Among other systems, protocol 2 was the more challenging protocol in the MFCC-SVM and SCFC-GMM baselines.

Finally, Table \ref{tab:AudioVisualPAD} reports the score mean fusion of Color Textures-SVM and LFCC-GMM baselines (best performing systems of each modality) on Audio-Visual attacks.
The low-quality attacks (PA.F.5 and PA.V.7) were detected with error rates less than $1\%$.
However, the performance is degraded significantly when high quality attacks are used with a D-EER of $29.34\%$ on the testing database.
This degradation in performance is also evident in protocol 2 where both low and high quality attacks are used.

\section{Research Potential for SWAN Multimodal Biometric Dataset}
\label{sec:PoT}
In this section, we summarize the research potential that can be anticipated using SWAN multimodal biometric dataset. As emphasized in Section \ref{Sec:DBUnique} that, the SWAN multimodal dataset includes new challenging scenarios and evaluation protocols that are tailored for the verification experiments.  This dataset is unique in terms of the number of data subjects captured in four different geographical locations using the same capture protocols. Thus, the following are research direction that can be pursued with SWAN multimodal biometric dataset:
\begin{itemize}
\item Developing novel algorithms for smartphone-based biometric verification on both unimodal (face, voice and periocular) and multimodal biometric characteristics.
\item Study of variation in developed biometric systems (both unimodal and multimodal) due to environmental variations captured in six different sessions.
\item Study on multi-lingual speaker verification as each data-subject has recorded its voice samples in multiple languages.
\item Evaluation of both unimodal and multimodal Presentation Attack Detection (PAD) algorithms.
\item Generation of new Presentation Attack Instrument (PAI) and also the vulnerability evaluation of potential PAIs.
\item Demographic study including age, gender, or visual aids.
\item Study on cross-European diversity and geographic variability in terms of  both unimodal and multimodal biometrics.
\end{itemize}
\section{Data-set distribution}
\label{sec:DBDis}
The Idiap subset of the database will be distributed online on the Idiap dataset distribution portal (\url{https://www.idiap.ch/dataset}). Also we are aiming to make the complete database available through the BEAT platform\footnote{\url{https://www.beat-eu.org/platform/}} \cite{beat2017}.
The BEAT platform is a European computing e-infrastructure solution for open access and scientific information sharing and re-use. Both the data and the source code of experiments are shared while protecting privacy and confidentiality.
As the data on the BEAT platform is easily available for experimental research but cannot be downloaded, it makes BEAT an ideal platform for biometrics research while the privacy of users who participated in the data collection is maintained.

\section{Conclusions}
\label{sec:Cocn}
Smartphone-based biometric verification has gained a lot of interest among researchers from the past few years. Availability of the publicly available datasets is crucial to driving the research forward so that new algorithms are developed and benchmarked with the existing algorithms. However, the collection of biometric datasets is resource consuming, especially in collecting the multimodal biometric dataset from different geographic locations. In this work, we present a new multimodal biometric dataset captured using a smartphone together with the evaluation of the baseline techniques.  This dataset is a result of the collaboration between three European partners within the framework of the SWAN project sponsored by Research Council of Norway.  This new dataset is captured in the challenging conditions in four different geographic locations. The whole dataset is obtained from 150 data subjects in six different sessions that can simulate the real-life scenarios. Besides, a new presentation attack (or spoofing) dataset is also presented for multimodal biometrics. A brief description of the performance evaluation protocols and the baseline biometric verification and presentation attack detection algorithms are benchmarked. Experimental findings using the baseline algorithms are highlighted for both biometric verification and presentation attack detection.

\section*{Acknowledgement}
This work is carried out under the  funding of the
Research Council of Norway (Grant No. IKTPLUSS
248030/O70)
{\tiny
	\bibliographystyle{IEEEtran}
	\bibliography{SWAN}

\begin{thebibliography}{10}
\providecommand{\url}[1]{#1}
\csname url@samestyle\endcsname
\providecommand{\newblock}{\relax}
\providecommand{\bibinfo}[2]{#2}
\providecommand{\BIBentrySTDinterwordspacing}{\spaceskip=0pt\relax}
\providecommand{\BIBentryALTinterwordstretchfactor}{4}
\providecommand{\BIBentryALTinterwordspacing}{\spaceskip=\fontdimen2\font plus
\BIBentryALTinterwordstretchfactor\fontdimen3\font minus
  \fontdimen4\font\relax}
\providecommand{\BIBforeignlanguage}[2]{{%
\expandafter\ifx\csname l@#1\endcsname\relax
\typeout{** WARNING: IEEEtran.bst: No hyphenation pattern has been}%
\typeout{** loaded for the language `#1'. Using the pattern for}%
\typeout{** the default language instead.}%
\else
\language=\csname l@#1\endcsname
\fi
#2}}
\providecommand{\BIBdecl}{\relax}
\BIBdecl

\bibitem{ISO-IEC-2382-37-170206}
{ISO/IEC JTC1 SC37 Biometrics}, \emph{{ISO/IEC} 2382-37:2017 Information
  Technology - Vocabulary - Part 37: Biometrics}, International Organization
  for Standardization, 2017.

\bibitem{AcuityWebPage}
``{Acuity Intelligence Forecast},'' \url{http://www.acuity-mi.com}, accessed:
  2019-04-29.

\bibitem{TrendWiseWebPage}
\BIBentryALTinterwordspacing
``\BIBforeignlanguage{english}{Global {Smartphone} {Production} {Volume} {May}
  {Decline} by {Up} to 5\% in 2019, {Huawei} {Would} {Overtake} {Apple} to
  {Become} {World}'s {Second} {Largest} {Smartphone} {Maker}, {Says}
  {TrendForce}}.'' [Online]. Available:
  \url{https://press.trendforce.com/node/view/3200.html}
\BIBentrySTDinterwordspacing

\bibitem{Ramachandra:2017:PAD:3058791.3038924}
\BIBentryALTinterwordspacing
R.~Ramachandra and C.~Busch, ``Presentation attack detection methods for face
  recognition systems: A comprehensive survey,'' \emph{ACM Comput. Surv.},
  vol.~50, no.~1, pp. 8:1--8:37, Mar. 2017. [Online]. Available:
  \url{http://doi.acm.org/10.1145/3038924}
\BIBentrySTDinterwordspacing

\bibitem{marcel2018handbook}
S.~Marcel, M.~S. Nixon, and S.~Z. Li, \emph{Handbook of biometric
  anti-spoofing}.\hskip 1em plus 0.5em minus 0.4em\relax Springer, 2018,
  vol.~1.

\bibitem{zhang2015fingerprints}
Y.~Zhang, Z.~Chen, H.~Xue, and T.~Wei, ``Fingerprints on mobile devices:
  Abusing and leaking,'' in \emph{Black Hat Conference}, 2015.

\bibitem{7893784}
A.~{Roy}, N.~{Memon}, and A.~{Ross}, ``Masterprint: Exploring the vulnerability
  of partial fingerprint-based authentication systems,'' \emph{IEEE
  Transactions on Information Forensics and Security}, vol.~12, no.~9, pp.
  2013--2025, Sep. 2017.

\bibitem{RATTANI201839}
\BIBentryALTinterwordspacing
A.~Rattani and R.~Derakhshani, ``A survey of mobile face biometrics,''
  \emph{Computers \& Electrical Engineering}, vol.~72, pp. 39 -- 52, 2018.
  [Online]. Available:
  \url{http://www.sciencedirect.com/science/article/pii/S004579061730650X}
\BIBentrySTDinterwordspacing

\bibitem{MICHEI}
\BIBentryALTinterwordspacing
M.~De~Marsico, M.~Nappi, D.~Riccio, and H.~Wechsler, ``Mobile iris challenge
  evaluation (miche)-i, biometric iris dataset and protocols,'' \emph{Pattern
  Recogn. Lett.}, vol.~57, no.~C, pp. 17--23, May 2015. [Online]. Available:
  \url{http://dx.doi.org/10.1016/j.patrec.2015.02.009}
\BIBentrySTDinterwordspacing

\bibitem{Patel}
V.~M. {Patel}, R.~{Chellappa}, D.~{Chandra}, and B.~{Barbello}, ``Continuous
  user authentication on mobile devices: Recent progress and remaining
  challenges,'' \emph{IEEE Signal Processing Magazine}, vol.~33, no.~4, pp.
  49--61, July 2016.

\bibitem{FingerPhoto}
A.~{Sankaran}, A.~{Malhotra}, A.~{Mittal}, M.~{Vatsa}, and R.~{Singh}, ``On
  smartphone camera based fingerphoto authentication,'' in \emph{2015 IEEE 7th
  International Conference on Biometrics Theory, Applications and Systems
  (BTAS)}, Sep. 2015, pp. 1--7.

\bibitem{6266494}
C.~{McCool}, S.~{Marcel}, A.~{Hadid}, M.~{Pietikäinen}, P.~{Matejka},
  J.~{Cernocký}, N.~{Poh}, J.~{Kittler}, A.~{Larcher}, C.~{Lévy},
  D.~{Matrouf}, J.~{Bonastre}, P.~{Tresadern}, and T.~{Cootes}, ``Bi-modal
  person recognition on a mobile phone: Using mobile phone data,'' in
  \emph{2012 IEEE International Conference on Multimedia and Expo Workshops},
  July 2012, pp. 635--640.

\bibitem{bartuzi2018mobibits}
E.~Bartuzi, K.~Roszczewska, R.~Bia{\l}obrzeski \emph{et~al.}, ``Mobibits:
  Multimodal mobile biometric database,'' in \emph{2018 International
  Conference of the Biometrics Special Interest Group (BIOSIG)}.\hskip 1em plus
  0.5em minus 0.4em\relax IEEE, 2018, pp. 1--5.

\bibitem{SWANPage}
``{Secure Access Control over Wide Area Network (SWAN)},''
  \url{https://www.ntnu.edu/iik/swan/}, accessed: 2019-08-09.

\bibitem{CSIAP_DB}
\BIBentryALTinterwordspacing
G.~Santos, E.~Grancho, M.~V. Bernardo, and P.~T. Fiadeiro, ``Fusing iris and
  periocular information for cross-sensor recognition,'' \emph{Pattern
  Recognition Letters}, vol.~57, pp. 52 -- 59, 2015, mobile Iris CHallenge
  Evaluation part I (MICHE I). [Online]. Available:
  \url{http://www.sciencedirect.com/science/article/pii/S0167865514003006}
\BIBentrySTDinterwordspacing

\bibitem{FTV_DB}
D.~{Kim}, K.~{Chung}, and K.~{Hong}, ``Person authentication using face, teeth
  and voice modalities for mobile device security,'' \emph{IEEE Transactions on
  Consumer Electronics}, vol.~56, no.~4, pp. 2678--2685, November 2010.

\bibitem{MobBIO}
A.~F. {Sequeira}, J.~C. {Monteiro}, A.~{Rebelo}, and H.~P. {Oliveira},
  ``Mobbio: A multimodal database captured with a portable handheld device,''
  in \emph{2014 International Conference on Computer Vision Theory and
  Applications (VISAPP)}, vol.~3, Jan 2014, pp. 133--139.

\bibitem{UMDAA}
U.~{Mahbub}, S.~{Sarkar}, V.~M. {Patel}, and R.~{Chellappa}, ``Active user
  authentication for smartphones: A challenge data set and benchmark results,''
  in \emph{2016 IEEE 8th International Conference on Biometrics Theory,
  Applications and Systems (BTAS)}, Sep. 2016, pp. 1--8.

\bibitem{MobiBits_DB}
E.~{Bartuzi}, K.~{Roszczewska}, M.~{rokielewicz}, and R.~{Białobrzeski},
  ``Mobibits: Multimodal mobile biometric database,'' in \emph{2018
  International Conference of the Biometrics Special Interest Group (BIOSIG)},
  Sep. 2018, pp. 1--5.

\bibitem{BioSecure_DB}
J.~{Ortega-Garcia}, J.~{Fierrez}, F.~{Alonso-Fernandez}, J.~{Galbally}, M.~R.
  {Freire}, J.~{Gonzalez-Rodriguez}, C.~{Garcia-Mateo}, J.~{Alba-Castro},
  E.~{Gonzalez-Agulla}, E.~{Otero-Muras}, S.~{Garcia-Salicetti}, L.~{Allano},
  B.~{Ly-Van}, B.~{Dorizzi}, J.~{Kittler}, T.~{Bourlai}, N.~{Poh}, F.~{Deravi},
  M.~N.~R. {Ng}, M.~{Fairhurst}, J.~{Hennebert}, A.~{Humm}, M.~{Tistarelli},
  L.~{Brodo}, J.~{Richiardi}, A.~{Drygajlo}, H.~{Ganster}, F.~M. {Sukno},
  S.~{Pavani}, A.~{Frangi}, L.~{Akarun}, and A.~{Savran}, ``The multiscenario
  multienvironment biosecure multimodal database (bmdb),'' \emph{IEEE
  Transactions on Pattern Analysis and Machine Intelligence}, vol.~32, no.~6,
  pp. 1097--1111, June 2010.

\bibitem{parkhi_deep_2015}
O.~M. Parkhi, A.~Vedaldi, and A.~Zisserman, ``Deep face recognition,'' in
  \emph{British {{Machine Vision Conference}}}, vol.~1.\hskip 1em plus 0.5em
  minus 0.4em\relax BMVA Press, 09 2015, pp. 41.1 -- 41.12.

\bibitem{schroff_facenet:_2015}
F.~Schroff, D.~Kalenichenko, and J.~Philbin, ``{FaceNet}: {{A}} {U}nified
  {E}mbedding for {F}ace {R}ecognition and {C}lustering,'' in \emph{Proc. IEEE
  Conf. on Computer Vision and Pattern Recognition (CVPR)}.\hskip 1em plus
  0.5em minus 0.4em\relax IEEE, 2015, pp. 815 -- 823, 00297.

\bibitem{sandberg_facenet_2017}
D.~Sandberg, ``facenet: Face recognition using tensorflow,''
  \url{https://github.com/davidsandberg/facenet}, accessed: 2017-08-01.

\bibitem{LFWTech}
G.~B. Huang, M.~Ramesh, T.~Berg, and E.~Learned-Miller, ``{Labeled Faces in the
  Wild: A Database for Studying Face Recognition in Unconstrained
  Environments},'' University of Massachusetts, Amherst, Tech. Rep. 07-49,
  October 2007.

\bibitem{ICIPCompitRaghu}
R.~{Raghavendra} and C.~{Busch}, ``Learning deeply coupled autoencoders for
  smartphone based robust periocular verification,'' in \emph{2016 IEEE
  International Conference on Image Processing (ICIP)}, Sep. 2016, pp.
  325--329.

\bibitem{ICIPCompitKiran}
K.~B. {Raja}, R.~{Raghavendra}, and C.~{Busch}, ``Collaborative representation
  of deep sparse filtered features for robust verification of smartphone
  periocular images,'' in \emph{2016 IEEE International Conference on Image
  Processing (ICIP)}, Sep. 2016, pp. 330--334.

\bibitem{VOGT200817}
\BIBentryALTinterwordspacing
R.~Vogt and S.~Sridharan, ``Explicit modelling of session variability for
  speaker verification,'' \emph{Computer Speech \& Language}, vol.~22, no.~1,
  pp. 17 -- 38, 2008. [Online]. Available:
  \url{http://www.sciencedirect.com/science/article/pii/S0885230807000277}
\BIBentrySTDinterwordspacing

\bibitem{McCool_IET_BMT_2013}
C.~McCool, R.~Wallace, M.~McLaren, L.~El~Shafey, and S.~Marcel, ``Session
  variability modelling for face authentication,'' \emph{IET Biometrics},
  vol.~2, no.~3, pp. 117--129, Sep. 2013.

\bibitem{le_robust_2018}
N.~Le and J.-M. Odobez, ``Robust and discriminative speaker embedding via
  intra-class distance variance regularization,'' in \emph{Proceedings
  {Interspeech}}, 2018, pp. 2257--2261.

\bibitem{reynolds_speaker_2000}
D.~A. Reynolds, T.~F. Quatieri, and R.~B. Dunn, ``Speaker verification using
  adapted {Gaussian} mixture models,'' \emph{Digital signal processing},
  vol.~10, no.~1, pp. 19--41, 2000.

\bibitem{nagrani_voxceleb:_2017}
\BIBentryALTinterwordspacing
A.~Nagrani, J.~S. Chung, and A.~Zisserman, ``{VoxCeleb}: a large-scale speaker
  identification dataset,'' \emph{arXiv:1706.08612 [cs]}, Jun. 2017, arXiv:
  1706.08612. [Online]. Available: \url{http://arxiv.org/abs/1706.08612}
\BIBentrySTDinterwordspacing

\bibitem{BioEvaSpoof}
I.~Chingovska, A.~AndrÃ©, and S.~Marcel, ``Biometrics evaluation under
  spoofing attacks,'' \emph{IEEE Transactions on Information Forensics and
  Security (T-IFS)}, vol.~9, no.~12, pp. 2264--2276, Dec 2014.

\bibitem{Raghusipco}
R.~Ramachandra and C.~Busch, ``Presentation attack detection algorithm for face
  and iris biometrics,'' in \emph{22nd European Signal Processing Conference,
  {EUSIPCO} 2014, Lisbon, Portugal}, 2014, pp. 1387--1391.

\bibitem{IDMJain}
D.~Wen, H.~Han, and A.~Jain, ``Face spoof detection with image distortion
  analysis,'' \emph{IEEE Transactions on Information Forensics and Security},
  vol.~10, no.~99, pp. 1--16, 2015.

\bibitem{ColorLBPfacePAD}
Z.~Boulkenafet, J.~Komulainen, and A.~Hadid, ``Face anti-spoofing based on
  color texture analysis,'' in \emph{IEEE International Conference on Image
  Processing (ICIP)}.\hskip 1em plus 0.5em minus 0.4em\relax IEEE, 2015, pp.
  2636--2640.

\bibitem{sahidullah_comparison_2015}
\BIBentryALTinterwordspacing
M.~Sahidullah, T.~Kinnunen, and C.~Hanilçi, ``A comparison of features for
  synthetic speech detection,'' in \emph{Proc. {INTERSPEECH}}.\hskip 1em plus
  0.5em minus 0.4em\relax Citeseer, 2015, pp. 2087--2091. [Online]. Available:
  \url{http://citeseerx.ist.psu.edu/viewdoc/download?doi=10.1.1.709.5379&rep=rep1&type=pdf}
\BIBentrySTDinterwordspacing

\bibitem{ISO/IEC2015a}
{ISO/IEC JTC1 SC37 Biometrics}, \emph{{ISO/IEC} 30107-3. Information Technology
  - Biometric presentation attack detection - Part 3: Testing and Reporting},
  International Organization for Standardization, 2017.

\bibitem{VGG}
O.~M. Parkhi, A.~Vedaldi, and A.~Zisserman, ``Deep face recognition,'' in
  \emph{British Machine Vision Conference}, 2015.

\bibitem{beat2017}
\BIBentryALTinterwordspacing
A.~Anjos, L.~El-Shafey, and S.~Marcel, ``Beat: An open-science web platform,''
  in \emph{International Conference on Machine Learning (ICML)}, Aug. 2017.
  [Online]. Available:
  \url{https://publications.idiap.ch/downloads/papers/2017/Anjos_ICML2017_2017.pdf}
\BIBentrySTDinterwordspacing

\end{thebibliography}
}

\end{document}